\documentclass[lettersize,journal]{IEEEtran}

\usepackage{amsmath,amsfonts,amssymb}
\usepackage{algorithm}
\usepackage{algorithmic}
\usepackage{array}
\usepackage[caption=false,font=footnotesize]{subfig}
\usepackage{textcomp}
\usepackage{stfloats}
\usepackage{url}
\usepackage{verbatim}
\usepackage{graphicx}
\usepackage{cite}
\usepackage{booktabs}
\usepackage{multirow}
\usepackage[table]{xcolor}
\usepackage{balance}
\usepackage[hidelinks]{hyperref}

\usepackage{tabularx}

\newcommand{\method}{R$^2$LPL}
\newcommand{\rocl}{ROCL}
\newcommand{\state}{s}
\newcommand{\action}{a}

\newcommand{\dataset}{\mathcal{D}}
\newcommand{\memory}{\mathcal{M}}
\newcommand{\anchors}{\mathcal{A}}
\newcommand{\traj}{\tau}

\begin{document}

\title{Learning from Mistakes: Rollout-Retrieval Lifelong Policy Learning for Autonomous Driving}

\author{Cheng Gong, Haoyang Wang, Chao Lu, Zirui Li, Jianwei Gong
\thanks{This work was supported in part by the National Natural Science Foundation of China under Grant 52372405, and in part by the National Key R\&D Program of China under Grant 2022ZD0115503. Corresponding authors: Chao Lu and Jianwei Gong.

Cheng Gong, Haoyang Wang, Chao Lu, and Jianwei Gong are with the School of Mechanical Engineering, Beijing Institute of Technology, Beijing 100081, China (e-mails: engibacter@gmail.com; haoyaw7@gmail.com; chaolu@bit.edu.cn; gongjianwei@bit.edu.cn).

Zirui Li is with the School of Mechanical and Aerospace Engineering, Nanyang Technological University, Singapore 639798 (e-mail: zirui.li@ntu.edu.sg).}
}

\markboth{Preprint}%
{Gong \MakeLowercase{\textit{et al.}}: Rollout-Retrieval Lifelong Policy Learning}

\maketitle

\begin{abstract}
Autonomous driving policies should be able to improve continually as deployment exposes them to increasingly diverse and long-tail traffic situations. However, most learning-based policies are trained or fine-tuned on expert demonstrations and then rely largely on generalization to handle challenging closed-loop scenarios, lacking an explicit mechanism to correct and retain the mistakes exposed in these scenarios. This paper studies autonomous driving policy improvement from a lifelong learning perspective: Can a pretrained policy improve continually by accumulating corrective knowledge derived from its own mistakes, while retaining previously acquired driving competence? To answer this question, we propose \emph{Rollout-Retrieval Lifelong Policy Learning} (\method), a new policy learning framework that retrieves corrective targets from recoverable policy-induced mistakes and retains the resulting knowledge through lifelong policy learning. \method\ addresses a key bottleneck in continual policy improvement: closed-loop mistakes reveal where the policy is weak, but do not directly specify what the policy should learn. By filtering recoverable mistake-related states and retrieving feasible corrective targets, \method\ turns sparse failure evidence into compact supervised knowledge for stable and sample-efficient policy improvement. We evaluate \method\ on large-scale closed-loop nuPlan benchmarks. With only a few rollout and continual-learning cycles, \method\ elevates a learning-based planner with moderate initial performance to state-of-the-art performance across the evaluated benchmarks, especially on the challenging and long-tail \texttt{Test14-hard} split. These results demonstrate the effectiveness of \method\ in converting recoverable closed-loop mistakes into corrective knowledge for sustained policy improvement.
Code available at: \url{https://github.com/Engibacter/R2LPL}.
\end{abstract}

\begin{IEEEkeywords}
Autonomous driving, motion planning, lifelong learning, continual learning, policy learning, long-tail scenarios, closed-loop decision making.
\end{IEEEkeywords}

\section{Introduction}
\IEEEPARstart{A}{utonomous} vehicles are expected to operate in open and continually evolving environments, where rare combinations of road geometry, traffic interactions, and agent behaviors inevitably give rise to long-tail situations \cite{Karnchanachari2024NuplanBenchmark,Dauner2024NAVSIM}. Recent advances in planning-oriented and end-to-end autonomous driving have substantially improved closed-loop performance \cite{Hu2023UniAD,Chitta2023Transfuser,e2esurvey2024chen}  through increasingly powerful policy architectures \cite{cheng2024pluto, sun2025sparsedrive, hydramdp2024li}, 
generative models \cite{diffusiondrive2025,diffusionplanner2025zheng,meanfuser2026wang}, 
and large-scale foundation models \cite{OpenDriveVLA2025, AutoVLA2025, ReasoningVLA2025}.
Despite these advances, mistakes remain inevitable. Even highly capable driving policies may encounter situations in which their learned behaviors are insufficient, leading to suboptimal decisions or closed-loop failures \cite{dagger2011ross, RoaD2026garcia}. These failures may stem from rare interactions, imperfect generalization, or the accumulation of policy-induced deviations over closed-loop execution. This raises a fundamental question: if mistakes are inevitable, can a deployed driving policy learn from them and continually improve itself over time?

Most existing learning-based planning policies address this challenge by exploiting richer information from pre-collected expert data. Imitation-based planners improve scene representations, model architectures, interaction reasoning, and training strategies to obtain stronger generalization from expert demonstrations~\cite{Huang_2023_ICCV,cheng2023plantf,cheng2024pluto,chen2024vadv2,hydramdp2024li}. Generative planners further model multimodal trajectory distributions using diffusion- or flow-based formulations~\cite{diffusionplanner2025zheng,tan2025flow,Zhang2026DFP}. Vision-language-action and language-augmented driving models introduce high-level reasoning and broader semantic priors to improve adaptability in complex scenarios~\cite{OpenDriveVLA2025,AutoVLA2025,ReasoningVLA2025,Zheng2026PlanAgent}. 
These approaches have significantly advanced learning-based planning through increasingly powerful policy models and stronger generalization from pre-collected expert data. However, they primarily focus on better utilizing existing knowledge, offering limited mechanisms to incorporate new knowledge revealed by policy failures and difficult scenarios.

A promising direction is to continuously acquire new knowledge from policy interactions and exploit the resulting feedback to further improve policy behavior. 
Reinforcement learning and preference-based optimization can improve closed-loop behavior beyond expert demonstration, but reward design, sparse safety feedback, and exploration risk remain major challenges for safety-critical driving \cite{Wu2023HG-RL,zhang2025carplanner,Shang2025DriveDPO,tang2026planr1,DenseLearning2026}. 
Dataset aggregation methods such as DAgger~\cite{dagger2011ross} improve policies by rolling out the current learner and querying an expert on visited states. However, obtaining expert supervision for arbitrary policy-induced states is often prohibitively expensive in large-scale driving scenarios. Recent closed-loop supervised fine-tuning methods, such as CAT-K~\cite{catk2025zhang} and RoaD~\cite{RoaD2026garcia}, alleviate this burden by using logged ground-truth trajectories to guide closed-loop data generation and then fine-tuning driving policies on the recovered rollout experiences.
Nevertheless, the knowledge recovered from closed-loop rollout is often heterogeneous in quality and informativeness. As policies deviate from expert states in rollout, some recovered knowledge provide valuable corrective information, whereas others may introduce biased or misleading updates. Consequently, effectively extracting useful corrective knowledge from closed-loop rollout remains an open challenge.

\begin{figure}[!t]
\centering
\includegraphics[width=\columnwidth]{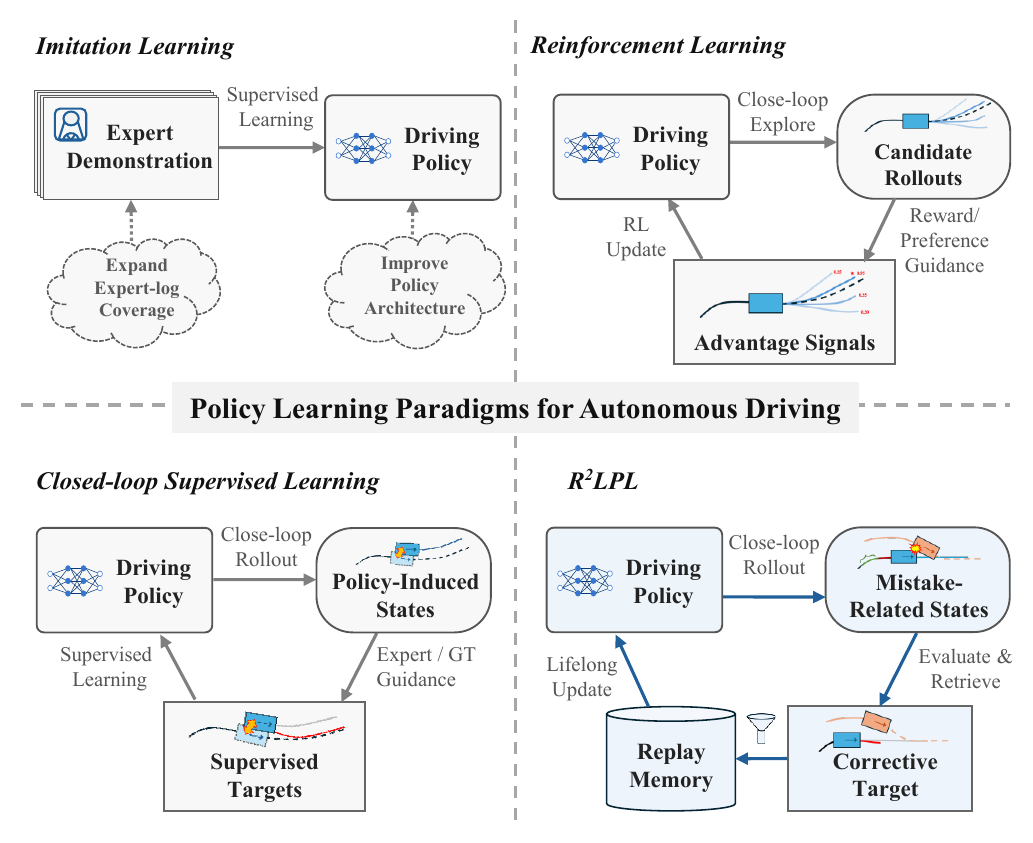}
\caption{Comparison of policy learning paradigms for autonomous driving. Existing paradigms improve policies from fixed expert demonstrations, reward/advantage signals, or expert/ground-truth guided supervision on policy-induced rollouts. In contrast, the proposed R$^2$LPL converts recoverable policy-induced mistakes into corrective targets and accumulates the resulting knowledge through replay memory, providing a lifelong policy improvement route driven by the policy's own mistakes.}
\label{fig:teaser}
\end{figure}

We study this problem from a lifelong policy learning perspective. As shown in Fig.~\ref{fig:teaser}, in contrast to imitation learning, which relies on stronger generalization from fixed expert demonstrations, and reinforcement learning, which improves policies through reward-driven trial and error, our goal is to enable a pretrained planner to improve through repeated correction of its own rollout-induced mistakes. Unlike dataset aggregation, this process does not assume that a queryable expert can provide valid labels for arbitrary learner-induced states. Instead, it treats closed-loop rollouts as a stream of policy-dependent experience, from which the learner must selectively retrieve recoverable mistakes and convert them into reliable corrective knowledge.
This perspective raises two central challenges. The first is \emph{mistake-to-knowledge conversion}: how to accurately and efficiently extract reliable corrective knowledge from heterogeneous rollout experience. Potential policy mistakes may lead to failures, risks, and behavioral conflicts events in close-loop rollout. Yet such outcomes do not directly reveal the responsible mistakes or the supervision targets that should be retrieved. Since only part of the induced experience contains actionable corrective information, the learner must identify informative mistake-related states and retrieve valid targets while avoiding ambiguous or noisy supervision.
The second is \emph{non-forgetting policy improvement}: how to improve over repeated correction cycles without forgetting previous corrections. Once a policy is updated, it induces a new rollout distribution and may expose new mistakes. The challenge is to learn from these newly revealed mistakes while retaining the knowledge that prevents old mistakes from reappearing.

To tackle these challenges, this paper proposes \method, short for \emph{Rollout-Retrieval Lifelong Policy Learning}, as a paradigm for lifelong policy improvement. \method\ first uses closed-loop rollouts to collect policy-induced experience associated with failures, risks, and conflicts. To address mistake-to-knowledge conversion, \method\ performs rollout-retrieval (R$^2$): candidate actions are evaluated at selected policy-induced states, feasible targets are retrieved from states that still admit meaningful correction, and misleading targets are avoided when no feasible correction exists. To address non-forgetting policy improvement, \method\ further performs lifelong policy learning (LPL), which updates the planner with newly retrieved R$^2$ knowledge while replaying previously learned corrections. Through repeated rollout-continual-learning (\rocl) cycles, \method\ enables a pretrained planner to continuously improve from its own mistakes without relying on reward-driven exploration or online expert querying.

The main contributions of this paper are:
\begin{itemize}
    \item We introduce \method, a rollout-retrieval lifelong policy learning framework for autonomous driving. It formulates policy learning and improvement as a process of learning from the policy's own mistakes, and organizes this process into mistake-related state mining, recoverability-aware corrective-knowledge construction, and lifelong policy update.
    \item We propose R$^2$, a rollout-retrieval mechanism that converts closed-loop mistakes into corrective supervision. Instead of directly using all rollout experiences, R$^2$ selectively mines mistake-related but recoverable states and retrieves feasible corrective targets from candidate actions.
    \item We formulate lifelong policy learning for planning as incremental accumulation of corrective knowledge. Instead of performing one-shot supervised fine-tuning on newly retrieved corrections, \method\ updates the policy with both new R$^2$ knowledge and replayed historical corrections, enabling multi-round improvement while reducing forgetting of previously corrective knowledge.
    \item We validate \method\ on large-scale closed-loop nuPlan benchmarks, showing that a learning-based planner with moderate initial performance can be improved to state-of-the-art performance within only a few \rocl\ rounds, especially on the hard long-tail \texttt{Test14-hard} split.

\end{itemize}

\section{Related Work}

\subsection{Planning Policy Learning for Autonomous Driving}

In planning policy learning, autonomous-driving planners are commonly formulated as parameterized policies that map scene observations to future ego trajectories or distributions over candidate trajectories. 
Early workers utilize transformer-based, vectorized, and interaction-aware modeling to encode driving scene, and regress or classify to the logged ego future trajectory \cite{mp32021casas,vad2023jiang}. GameFormer~\cite{Huang_2023_ICCV} formulates interactive prediction and planning with a game-theoretic transformer, while PLUTO~\cite{cheng2024pluto} advances imitation-based planning through strong scene representations and data augmentation. Another line of work formulates planning as  selection or scoring over multiple candidate actions. VADv2~\cite{chen2024vadv2} formulates end-to-end planning as a probabilistic distribution over a fixed set of trajectory anchors. Hydra-MDP~\cite{hydramdp2024li} uses multi-target rule-distillation to learn multi-modal planning targets. Recent mixture-of-experts planning improves generalization by composing specialized motion-planning behaviors \cite{Sun2025GeneralizingMotionPlanners}. Compared with black-box trajectory regression, probabilistic or score-based planners expose alternative action hypotheses and their relative preferences, which provides a structured interface for analyzing and improving policy decisions at closed-loop induced states.

Recent studies further improve learned planners through generative modeling and language-augmented policy architectures. Generative planners, including DiffusionDrive~\cite{diffusiondrive2025}, Diffusion Planner~\cite{diffusionplanner2025zheng}, Flow Planner~\cite{tan2025flow}, Diffusion Forcing Planner~\cite{Zhang2026DFP}, and MISTY~\cite{Xing2026MISTY}, model multimodal and uncertain driving behaviors with diffusion, flow matching, or other distributional trajectory-generation mechanisms. Meanwhile, vision-language-action and language-augmented models such as OpenDriveVLA~\cite{OpenDriveVLA2025}, AutoVLA~\cite{AutoVLA2025}, and Reasoning-VLA~\cite{ReasoningVLA2025} introduce semantic reasoning and cross-scenario priors into end-to-end driving policies. These methods demonstrate the importance of scaling policy representations, trajectory generation, and reasoning capability for autonomous driving. However, stronger policy architectures do not by themselves provide a mechanism for stable and continual policy improvement from closed-loop evaluation experience. In particular, the increasing complexity of generative objectives, guidance terms, large reasoning models, and end-to-end action generation makes targeted policy refinement sensitive to data distribution, optimization stability, inference efficiency, and safety verification.

\subsection{Closed-loop Policy Improvement}

Closed-loop policy improvement aims to refine a learned policy after initial training, especially when rollout execution exposes states that are not well handled by the pretrained policy. Classical dataset aggregation methods address this issue by learning from learner-induced states: DAgger~\cite{dagger2011ross} rolls out the learner, queries an expert at visited states, and aggregates newly labeled data, while DART~\cite{laskey2017dart}, SafeDAgger~\cite{zhang2016safedagger}, HG-DAgger~\cite{kelly2018hgdagger}, and EnsembleDAgger~\cite{menda2018ensembledagger} improve robustness, query efficiency, or safety through noise injection, safety classifiers, human intervention, or uncertainty estimation. These methods show the value of policy-induced states, but rely on expert or intervention signals that can provide valid supervision during rollout, which is difficult to obtain for large-scale autonomous-driving scenarios. Recent driving-oriented methods adapt closed-loop improvement to modern learned planners: CAT-K~\cite{catk2025zhang} performs closed-loop supervised fine-tuning for tokenized traffic models, and RoaD~\cite{RoaD2026garcia} uses expert-guided rollouts as demonstrations for supervised fine-tuning. While these works show that rollout experience can reveal weaknesses beyond fixed expert logs, such experience is not uniformly informative: policy-induced states may deviate from logged behaviors, differ in recoverability, or provide noisy correction signals. Moreover, each policy update changes the future rollout distribution, so policy improvement requires more than one-shot adaptation. It requires repeatedly acquiring reliable corrective knowledge while retaining previously learned corrections.

Reinforcement learning and preference-based optimization provide another route for policy improvement. Human-guided RL incorporates human knowledge to improve navigation behavior \cite{Wu2023HG-RL}; CarPlanner~\cite{zhang2025carplanner} applies large-scale reinforcement learning to autoregressive trajectory planning; DriveDPO~\cite{Shang2025DriveDPO} uses safety-oriented direct preference optimization for end-to-end driving; and Plan-R1~\cite{tang2026planr1} further represents trajectories as motion tokens and applies rule-based GRPO to align planning behavior with driving objectives. Recent diffusion-based planners, including DIVER~\cite{song2026diver-rl} and Hyper Diffusion Planner~\cite{zheng2026DiffusionRL}, further integrate reinforcement learning or reward optimization with generative trajectory modeling.These methods move beyond expert-log imitation and optimize policies toward closed-loop driving objectives. However, autonomous-driving rewards are sparse, multi-objective, and safety-critical, making exploration-intensive updates costly or unstable. Together with closed-loop supervised fine-tuning methods, these studies show that rollout experience is valuable for improving learned driving policies, but also highlight a central challenge: how to convert policy-induced experience into reliable and sustained policy improvement without depending on dense expert annotation or unstable exploration.

\subsection{Lifelong Policy Learning}

Lifelong learning, often referred to as continual learning, studies how a model can acquire knowledge from sequential or incremental data while mitigating catastrophic forgetting~\cite{clreview2022nmi,Wang2024ContinualLearningSurvey}. Representative strategies include regularization-based methods~\cite{Kirkpatrick2017EWC}, distillation-based methods~\cite{Li2016LwF}, replay-based methods~\cite{Rebuffi2017iCaRL,LopezPaz2017GEM,Buzzega2020DER}, and architecture- or parameter-isolation methods~\cite{DEN2018Yoon,PackNet2018Mallya}. With large pretrained models, recent studies further explore prompt- and parameter-efficient continual learning~\cite{Wang2022L2P,Wang2022DualPrompt,inflora2024liang,ptmreview2024ijcai}. These methods mainly address how to preserve past knowledge while adapting to a new sequence of given tasks or data distributions. Policy learning poses an additional difficulty: the data distribution is induced by the policy itself, and each policy update can change the states that will be encountered in future rollouts. Moreover, policy-induced states usually do not come with explicit ground-truth actions, so useful supervision must first be constructed from closed-loop experience. Therefore, lifelong policy learning requires not only non-forgetting optimization, but also a mechanism for acquiring reliable corrective knowledge from the evolving behavior of the policy.

Continual learning has also been explored in autonomous driving, including perception, prediction, and decision-making modules \cite{cdlreview2023li}. Prediction-oriented studies investigate case-level forgetting, dynamic expansion, and task-free identification for driving scene \cite{Li2026ContinualLearningInteractiveTrajectoryPrediction,Lin2026H2CPrediction}. 
These works extend continual learning to driving-specific data streams, but still largely follow the conventional setting where supervision from incremental data or task changes are given. 
Lifelong policy learning differs from conventional continual learning because the policy is updated through data that is partly induced by its own closed-loop behavior: its decisions affect future state visitation, and the resulting states may not come with explicit supervision.
Existing studies have explored continual policy learning from different angles, including dynamic confidence-aware reinforcement learning for autonomous driving \cite{Cao2023ContinuousImprovement}, lifelong skill preservation and recombination in robot manipulation \cite{Meng2025PreservingKnowledgeCombining}, life-long policy learning for path tracking control \cite{Gong2024BeyondImitationLifeLongPolicyLearning}, and human-guided continual learning for personalized driving decisions \cite{Yang2025HGCContinualLearning}.
These works show that policies can be improved over time through interaction, accumulated knowledge, or external feedback. However, they mainly focus on policy update or adaptation, while mistake-driven acquisition of corrective knowledge remains less explored. In closed-loop planning, failures do not directly reveal responsible decisions, recoverability, or corrective targets. Therefore, lifelong policy learning for planning requires not only preserving previous knowledge, but also converting mistake-related experience into reliable supervision for avoiding repeated failures.

\section{Problem Formulation}

\subsection{Planning Policy Learning for Autonomous Driving}
We consider autonomous-driving planning as trajectory-level policy learning. At time $t$, the planner observes a driving state
\begin{equation}
    \state_t =
    \{h_t^{\mathrm{ego}}, h_t^{\mathrm{agent}}, m_t, c_t, r_t\},
\end{equation}
where $h_t^{\mathrm{ego}}$ and $h_t^{\mathrm{agent}}$ denote ego and agent histories, $m_t$ denotes map and route geometry, $c_t$ denotes traffic-control context, and $r_t$ denotes the navigation goal. A learned planner maps $\state_t$ to a future ego trajectory $\traj_t$ or, more generally, to a distribution over candidate trajectories.

Because target retrieval requires candidate actions that can be explicitly searched, scored, and replayed as supervision, this paper focuses on score-based policies with a finite set of trajectory anchors as action space, similar to~\cite{chen2024vadv2,Shang2025DriveDPO}. Let
\begin{equation}
    \anchors=\{\action^1,\action^2,\ldots,\action^K\}
\end{equation}
be a finite anchor library, where each $\action^i=\{(x_\ell^i,y_\ell^i,\psi_\ell^i)\}_{\ell=1}^{H}$ is a future ego trajectory over horizon $H$. Given scene features $\phi(\state_t)$, the planner computes a trajectory score
\begin{equation}
    z_\theta^i(t) = f_\theta(\phi(\state_t), \action^i),
\end{equation}
where $\theta$ denotes the policy parameters and $z_\theta^i(t)$ is the logit associated with action $\action^i$ at state $\state_t$. The scores induce a categorical trajectory policy through softmax normalization:
\begin{equation}
    \pi_\theta(\action^i|\state_t)
    =
    \frac{\exp(z_\theta^i(t))}
    {\sum_{j=1}^{K}\exp(z_\theta^j(t))}.
    \label{eq:score_policy}
\end{equation}

The executed action can be obtained by maximum-score selection, sampling, or downstream rule-based selection among top candidates. Rather than treating the expert future only as a continuous regression target, the score-based planner is trained with sparse anchor-score supervision. For a driving state $\state$, let
\begin{equation}
    \mathcal{Y}(\state)=\{(i,y_i)\mid i\in\mathcal{A}_{\state}, y_i>0\}
\end{equation}
denote the supervised anchor-score set, where $\mathcal{A}_{\state}\subseteq{1,\ldots,K}$ is the subset of supervised anchor indices and $y_i$ is the non-negative target score assigned to anchor $\action^i$. Depending on the training source, $\mathcal{A}_{\state}$ and $y_i$ can be obtained from  scoring, rollout retrieval, or candidate sampling. The best supervised anchor is
\begin{equation}
i^*=\arg\max{(i,y_i)\in\mathcal{Y}(\state)}y_i .
\end{equation}
The normalized target preference over the supervised subset is defined as
\begin{equation}
q_i^{Y}(\state)=
\frac{y_i}{\sum_{j\in\mathcal{A}{\state}} y_j},
\quad i\in\mathcal{A}{\state}.
\label{eq}
\end{equation}
Correspondingly, we restrict the policy distribution to the same supervised subset:
\begin{equation}
q_{\theta}^{i}(\state)=
\frac{\exp(z_{\theta}^{i}(\state)/\epsilon)}
{\sum_{j\in\mathcal{A}{\state}}\exp(z{\theta}^{j}(\state)/\epsilon)},
\quad i\in\mathcal{A}_{\state},
\label{eq}
\end{equation}
where $\epsilon$ is a temperature parameter. This restricted distribution is used only for training efficiency with sparse supervision loss. During inference, the planner scores the full anchor library.

We define the anchor-score planning loss as
\begin{equation}
\begin{aligned}
\mathcal{L}_{\mathrm{AS}}(\state,\mathcal{Y};\theta)
=
-\log q{\theta}^{i^*}(\state) \
+
\lambda_{\mathrm{KL}}
\sum_{i\in\mathcal{A}{\state}}
q_i^{Y}(\state)
\log
\frac{q_i^{Y}(\state)}
{q{\theta}^{i}(\state)},
\end{aligned}
\label{eq:anchor_score_loss}
\end{equation}
where $\lambda_{\mathrm{KL}}$ balances the best-anchor classification term and the preference-matching term. The first term pulls the policy toward the best supervised anchor, while the second term preserves the relative preference among multiple positive anchors.

Standard pretraining can then be written as score-supervised planning over expert states,
\begin{equation}
    \dataset_E =
    \{(\state_t^E,\mathcal{Y}_t^E)\}_{t=1}^{N_E},
\end{equation}
where $\state_t^E$ is an expert-log state and $\mathcal{Y}_t^E$ is the corresponding sparse anchor-score supervision. The initial policy is obtained by: 
\begin{equation}
    \theta =
    \arg\min\theta
    \mathbb{E}_{(\state_t^E,\mathcal{Y}t^E)\sim\dataset_E}
    \left[
    \mathcal{L}_{\mathrm{AS}}
    \left(\state_t^E,\mathcal{Y}_t^E;\theta\right)
    \right]
    \label{eq:il_pretrain}
\end{equation}

During inference, the planner can score the full anchor library, whereas training may supervise only selected anchors for efficiency. This score-based formulation is important for rollout retrieval: the finite candidate action space exposes alternative decisions and allows a retrieved corrective target to be represented as replayable supervised knowledge at an arbitrary policy-induced state.

\subsection{Lifelong Policy Learning with Knowledge Replay}

Lifelong learning considers a sequence of learning tasks
$\{\dataset_1,\ldots,\dataset_K\}$, where $\dataset_k$ contains the supervised knowledge newly available at task $t$. Let $x$ denote a training sample, $\ell(x;\theta)$ its task loss under model parameters $\theta$, and $\theta^{k-1}$ and $\theta^k$ the model parameters before and after learning task $k$, respectively. The objective is to learn the new knowledge in $\dataset_k$ without overwriting useful knowledge acquired from $\dataset_1,\ldots,\dataset_{k-1}$.

In policy learning, the sequential tasks are not necessarily manually predefined. Instead, each task may correspond to a batch of incremental knowledge acquired from the current policy's closed-loop behavior. After the policy is updated, its future rollout distribution changes, and new weaknesses may be exposed. Thus, the knowledge available at each round is policy-dependent and can only be acquired progressively. A trivial alternative would be to store all knowledge collected from previous rounds and repeatedly train on their union. However, this corresponds to an unbounded joint-training setting that assumes unlimited memory and increasing optimization cost. Training only on the latest knowledge, on the other hand, may cause the policy to forget previously knowledge. We therefore formulate lifelong policy learning as bounded incremental policy improvement with knowledge replay. Let
\begin{equation}
    \memory_{k-1}
    \subseteq
    \bigcup_{j=1}^{k-1}\dataset_j,
    \qquad |\memory_{k-1}|\leq C,
    \label{eq:bounded_memory}
\end{equation}
denote the replay memory before round $k$, where $C$ is its maximum capacity. With replay mechanism, learning loss at task $k$ can be defined as:
\begin{equation}
\begin{split}
    \mathcal{L}_{\mathrm{replay}}^k(\theta)
    ={}&
    \mathbb{E}_{x\sim\dataset_k}
    \ell(x;\theta)\\
    &+
    \lambda_{\mathrm{rep}}
    \mathbb{E}_{x\sim\memory_{k-1}}
    \ell(x;\theta),
\end{split}
    \label{eq:cl_replay_objective}
\end{equation}
where the first term learns the current task, the second rehearses retained knowledge from historical tasks, and $\lambda_{\mathrm{rep}}\geq0$ controls their relative influence. 

Replaying only the supervised target preserves the expert-preferred feasible action, but ignores the expert or scoring knowledge about infeasible or less preferred alternatives. To retain both positive and negative preferences in the action space, we store the historical policy response and use it as an auxiliary distillation target during replay. Following the idea of dark-experience replay~\cite{Buzzega2020DER}, we store this response together with each replay sample. Let $u_x$ be the stored output vector for sample $x$, and let $o_\theta(x)$ be the current model output over the same dimensions. We define the response-retention loss as
\begin{equation}
    \mathcal{L}_{\mathrm{DER}}(x;\theta)
    =
    \mathrm{KL}
    \left(
    \mathrm{softmax}(u_x/\epsilon_d)
    \,\|\,
    \mathrm{softmax}(o_\theta(x)/\epsilon_d)
    \right),
    \label{eq:der_objective}
\end{equation}
where $\epsilon_d>0$ is a temperature parameter. Combining supervised replay with response retention gives
\begin{equation}
\begin{split}
    \mathcal{L}_{\mathrm{KR}}^k(\theta)
    ={}&
    \mathcal{L}_{\mathrm{replay}}^k(\theta)
    +
    \lambda_{\mathrm{DER}}
    \mathbb{E}_{x\sim\memory_{k-1}}
    \mathcal{L}_{\mathrm{DER}}(x;\theta),\\
    \theta^k={}&
    \arg\min_{\theta}\mathcal{L}_{\mathrm{KR}}^k(\theta),
\end{split}
    \label{eq:knowledge_replay_objective}
\end{equation}
where $\lambda_{\mathrm{DER}}\geq0$ controls retention of the historical model response. The replay term preserves stored supervision, while the DER term constrains changes in the richer output structure associated with that supervision.

After each task, the memory is updated by
\begin{equation}
    \memory_k
    =
    \mathcal{U}_C
    \left(\memory_{k-1}\cup\dataset_k\right),
    \label{eq:generic_memory_update}
\end{equation}
where $\mathcal{U}_C$ retains at most $C$ records from the previous memory and current data. 

\section{\method: Rollout-Retrieval Lifelong Policy Learning}

\subsection{Framework Overview}
\begin{figure*}[!t]
\centering
\includegraphics[width=\textwidth]{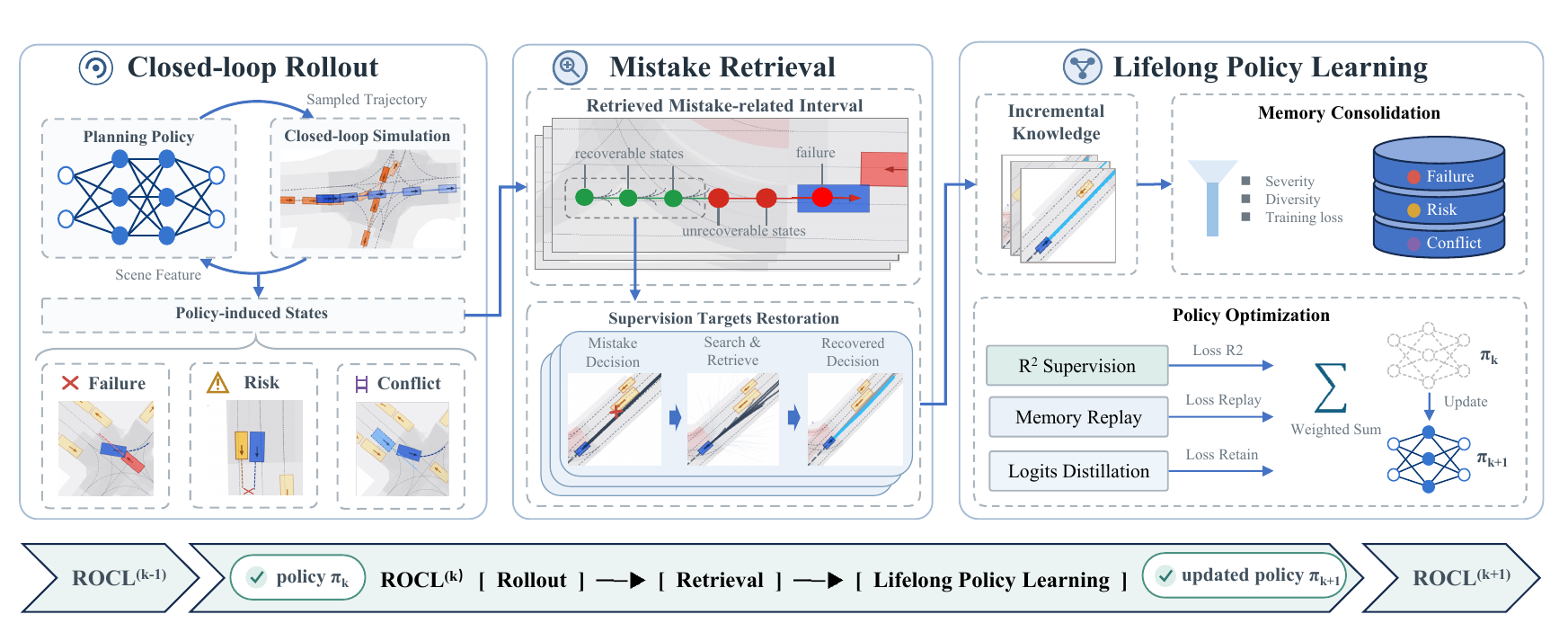}
\caption{Overview of \method. Each \rocl\ round follows a Rollout--Retrieval--Lifelong Policy Learning cycle: closed-loop rollout exposes failure, risk, and conflict evidence; retrieval identifies recoverable mistake-related states and constructs corrective targets; and lifelong policy learning updates the policy with new R$^2$ knowledge and replayed memory.}
\label{fig:overview}
\end{figure*}

Fig.~\ref{fig:overview} illustrates the overall pipeline of \method. Each \rocl\ round follows a Rollout--Retrieval--Lifelong Policy Learning cycle. The current policy is first executed in a closed-loop simulator or world model to induce states under its own decisions and expose failure, risk, or conflict evidence. The retrieval stage then searches the rollout experience for recoverable mistake-related states and converts them into corrective targets. Finally, lifelong policy learning updates the planner with the newly retrieved knowledge while replaying previous corrections. After the update, the improved policy induces a new closed-loop state distribution, which starts the next \rocl\ round.

The first two stages together instantiate rollout-retrieval (R$^2$) target construction. Rollout is used diagnostically rather than as a demonstration generator: it reveals where the current policy becomes unsafe, risky, or conflict-related under closed-loop execution. Retrieval further determines whether the mined mistake-related states are recoverable by the available trajectory-action space and selectively retrieves feasible candidate actions as corrective targets, while discarding unrecoverable states to avoid propagating misguiding rollout histories into supervision. The resulting R$^2$ knowledge is then used for lifelong policy learning, where the planner is updated with new R$^2$ supervision, replayed historical corrections, and anchor-score retention. Through repeated \rocl\ rounds, \method\ continually accumulates corrective knowledge from the policy's own closed-loop experience while reducing forgetting of previously learned corrective knowledge.

\subsection{R$^2$ Corrective Target Construction}

\subsubsection{Policy-Induced State Mining}
Static expert data describe states visited by the expert, but do not reveal which states the current policy will induce or which of its earlier decisions will lead to undesirable closed-loop behavior. We therefore roll out the current policy to obtain policy-dependent experience from which corrective knowledge can be constructed. At round $r$, executing $\pi_{r-1}$ in closed-loop rollout produces:
\begin{equation}
    \Xi_r
    =
    \operatorname{Rollout}(\pi_{r-1})
    =
    \{(\state_{r,t}^{\pi},\action_{r,t}^{\pi})\}_{t=1}^{T}.
    \label{eq:rollout_sequence}
\end{equation}
where $\Xi_r$ denotes the visited states collected from rollout in time $T$.

Not every visited state is informative for correction. We use three complementary event detectors
$d_g(\Xi_r,t)\in\{0,1\}$, $g\in\{\mathrm{F},\mathrm{R},\mathrm{C}\}$,
for Failure, Risk, and Conflict, respectively. Failure identifies states associated with an observed closed-loop failure; Risk captures safety-critical states that do not directly lead to failures; and Conflict captures inappropriate acceleration or yielding decisions that are inconsistent with logged expert behaviorally. Together, they cover terminal errors, non-terminal safety precursors, and decision-level inconsistencies.

Observed events are not necessarily caused by the action selected at the exact event frame: erroneous decisions can begin several steps earlier and gradually drive the policy into danger. We therefore assign each event a preceding credit-assignment window $t_{\mathrm{F}}$: 
\begin{equation}
    t_{\mathrm{F}}
    =
    \min\{t\mid d_{\mathrm{F}}(\Xi_r,t)=1\},
    \label{eq:failure_boundary}
\end{equation}

For event type $g$, a state is retained in $\mathcal{I}_{r,g}$, if a corresponding event occurs within the next $W_g$ steps:
\begin{equation}
    \mathcal{I}_{r,g}
    =
    \left\{
    \state_{r,t}^{\pi}
    \;\middle|\;
    \substack{
    t\leq t_{\mathrm{F}},\\
    \exists k\in[t,\min(t+W_g,t_{\mathrm{F}})]\\
    \text{s.t. }d_g(\Xi_r,k)=1
    }
    \right\}.
    \label{eq:event_candidate_indices}
\end{equation}

The potentially informative policy-induced states are therefore: 
\begin{equation}
    \mathcal{I}_r
    =
    \bigcup_{g\in\{\mathrm{F},\mathrm{R},\mathrm{C}\}}
    \mathcal{I}_{r,g}.
    \label{eq:mined_state_set}
\end{equation}

The horizon $W_g$ therefore links an observed event to the sequence of potentially responsible decisions rather than only supervising with its final frame. The constraint $t\leq t_{\mathrm{F}}$ excludes post-failure states whose histories have already been corrupted by mistake related actions. Applying this definition across rollouts yields the complete state set $\mathcal{I}_r$ for target retrieval.

\subsubsection{Action Scoring and Target Recovering}

State mining identifies where correction may be useful, but does not directly provide a valid target. In particular, the logged expert future is anchored at the logged state $\state_t^E$ and may become infeasible after the learned policy reaches a different state $\state_t^\pi$. R$^2$ therefore first estimates the reliability of log-based guidance and then searches the structured action space for a feasible correction.

For each $\state_{r,t}^{\pi}\in\mathcal{I}_r$, we omit the round superscript for clarity and characterize its deviation from the logged state by
\begin{equation}
    \Delta_t =
    \left(d_{xy}(\state_t^\pi,\state_t^E),
    d_\psi(\state_t^\pi,\state_t^E),
    d_v(\state_t^\pi,\state_t^E),
    d_p(\state_t^\pi,\state_t^E)\right)
    \label{eq:state_deviation}
\end{equation}

We first form a geometrically admissible subset $\anchors_t^{\mathrm{pre}}\subseteq\anchors$ by removing actions that are clearly incompatible with the route, heading, or drivable area while retaining stop actions. This preselection only reduces the number of actions evaluated. Each remaining action receives a rule-based planning score
\begin{equation}
    q_i
    =
    Q\!\left(
    m_i^{\mathrm{saf}},
    m_i^{\mathrm{drv}},
    m_i^{\mathrm{route}},
    m_i^{\mathrm{prog}},
    m_i^{\mathrm{comf}}
    \right),
    \label{eq:rule_planning_score}
\end{equation}
where the five terms measure safety, drivable-area compliance, route consistency, progress, and comfort, respectively, and $Q$ denotes their rule-based aggregation. The feasible candidate set $ \anchors_t^{\mathrm{valid}}$ and the state recoverability $R(\state_t^\pi)$ of state $\state_t^\pi$ are: 
\begin{equation}
\begin{split}
    \anchors_t^{\mathrm{valid}}
    &=
    \{\action_i\in\anchors_t^{\mathrm{pre}}\mid q_i>0\},\\
    R(\state_t^\pi)
    &=
    \mathbf{1}
    [\anchors_t^{\mathrm{valid}}\neq\emptyset].
\end{split}
    \label{eq:recoverability}
\end{equation}

States with $R(\state_t^\pi)=0$ are excluded because the available action space cannot provide valid supervision from them.

For a recoverable state, its deviation determines the guidance class $\kappa_t$:
\begin{equation}
    \kappa_t=\Gamma(\Delta_t)
    \in\{\mathrm{near},\mathrm{rec},\mathrm{far}\}.
    \label{eq:guidance_class}
\end{equation}
where $\mathrm{near}$, $\mathrm{rec}$, and $\mathrm{far}$ denote three types of guidance class. Near-log states retain reliable expert guidance; recoverable off-log states use the expert only as a weak behavioral prior; and far-off-policy states no longer use the logged expert future for target retrieval. Thus, recoverability determines whether the state can be supervised, whereas $\kappa_t$ determines which evidence should define that supervision.

For recoverable states, action quality is described by three signals: normalized rule score $\bar q_i$, reference consistency $c_i^{\mathrm{ref}}$, and expert consistency $c_i^E$. Reference consistency measures geometric alignment with the route/reference path. Expert consistency measures behavioral alignment with the logged future:
\begin{equation}
    \begin{aligned}
    c_i^{E}
    =
    \exp\big(
    &-\alpha_H d_H(\action_i,\traj_t^E)
    -\alpha_P d_P(\action_i,\traj_t^E) \\
    &-\alpha_\psi d_\psi(\action_i,\traj_t^E)
    -\alpha_v d_v(\action_i,\traj_t^E)
    \big),
    \end{aligned}
\end{equation}
where $d_H$ is a Hausdorff-style shape distance, $d_P$ measures progress mismatch along the reference path, and $d_\psi$ and $d_v$ measure heading and velocity disagreement, respectively. The progress term is symmetric: it penalizes both lagging and overshooting, so waiting and yielding behavior can be learned rather than overwritten by a progress-only scorer.

The final target score $y_i$ is computed as:
\begin{equation}
    y_i
    =
    w_q^{\kappa_t}\bar q_i
    +w_{\mathrm{ref}}^{\kappa_t}c_i^{\mathrm{ref}}
    +w_E^{\kappa_t}c_i^E .
    \label{eq:r2_score}
\end{equation}
where $w_q^{\kappa_t}$, $w_{\mathrm{ref}}^{\kappa_t}$ and $w_E^{\kappa_t}$ denote the respective weights for different guidance class $\kappa_t$.
The class-dependent parameters express the different reliability of supervision. Near-log states assign greater weight to expert consistency; recoverable off-log states emphasize rule and reference scores while retaining weak expert guidance; and far-off-policy states set $w_E^{\mathrm{far}}=0$. Hence the scoring rule transits from log-guided correction to rule-guided recovery as the deviation between policy-induced states and logged expert states increases. Thus the retrieved target is:
\begin{equation}
    \action^* = \arg\max_{\action_i \in \anchors_t^{\mathrm{valid}}} y_i.
\end{equation}

For training a score-based planner, the recovered sample stores the sparse target distribution: 
\begin{equation}
    \mathcal{Y}_t =
    \{(i,y_i)\mid \action_i\in\anchors_t^{\mathrm{valid}}, y_i>0\},
    \label{eq:r2_target}
\end{equation}
which supervises both the best anchor and the relative preference among recovered candidates.

\subsection{Lifelong Policy Learning with R$^2$ Knowledge}

Let $\dataset_r=\{(\state_t^\pi,\mathcal{Y}_t)\}$ denote the recovered knowledge produced by R$^2$ in round $r$. Each sample pairs a policy-induced driving state with a retrieved anchor-score target, so it can be used by the same score-based policy in Eq.~\eqref{eq:score_policy}. The key role of LPL is to incorporate this new R$^2$ knowledge for policy learning while retaining  knowledge learned in previous rounds. Thus both expert-log pretraining and rollout-induced lifelong learning are expressed through the same anchor-score planning loss in Eq.~\eqref{eq:anchor_score_loss}.

Let $\memory_r$ denote the bounded memory available before learning round $r$. The round-$r$ update minimizes: 
\begin{equation}
\begin{split}
    \mathcal{L}_r =
    &\mathbb{E}_{x\sim\dataset_r}
    \mathcal{L}_{\mathrm{AS}}(x;\theta)
    + \lambda_{\mathrm{rep}}
    \mathbb{E}_{x\sim\memory_r}
    \mathcal{L}_{\mathrm{AS}}(x;\theta) \\
    &+ \lambda_{\mathrm{DER}}
    \mathbb{E}_{x\sim\memory_r}
    \mathcal{L}_{\mathrm{DER}}(x;\theta)
    + \lambda_E
    \mathbb{E}_{x\sim\dataset_E^r}
    \mathcal{L}_{\mathrm{AS}}(x;\theta).
\end{split}
    \label{eq:lpl_instantiated}
\end{equation}
where $x$ is shorthand for a state--target pair $(\state,\mathcal{Y})$. The first term learns from the current recovered failures. The second term constraints policy from forgetting replayed knowledge from previous rounds. The DER term preserves the stored teacher preference over replayed anchors through score-distribution matching. The final term mixes original expert-log samples from $\dataset_E^r$ to retain broad log-distribution behavior.

After optimizing Eq.~\eqref{eq:lpl_instantiated}, LPL updates its bounded memory from both historical and newly recovered knowledge. The memory candidate pool is the duplicated union
\begin{equation}
    \mathcal{P}_r = \operatorname{Unique}
    \left(\memory_r\cup\dataset_r\right).
    \label{eq:memory_pool}
\end{equation}

As memory capacity is limited, it is important to update and store only the most valuable knowledge. Each $x\in\mathcal{P}_r$ is evaluated from two complementary aspects. The first is learning difficulty $ h_r(x)$. For a new sample, it is measured by the mean anchor-score loss observed over $N_x$ times it is sampled during round-$r$ training. For a historical sample that is not revisited as current data, its stored priority $ \rho_{r-1}(x)$ from the previous memory update is carried forward:
\begin{equation}
    h_r(x)=
    \begin{cases}
    \displaystyle\frac{1}{N_x}\sum_{n=1}^{N_x}
    \mathcal{L}_{\mathrm{AS}}^{(n)}(x;\theta),
    &x\in\dataset_r,\\[6pt]
    \rho_{r-1}(x),&x\in\memory_r\setminus\dataset_r.
    \end{cases}
    \label{eq:memory_difficulty}
\end{equation}

The second aspect is rollout utility $u(x)$, which favors samples associated with more consequential or more difficult-to-recover closed-loop states:
\begin{equation}
\begin{split}
    u(x)={}&
    \sum_{g\in\{\mathrm{F},\mathrm{R},\mathrm{C}\}}
    w_g\,\mathbf{1}[g\in\mathcal{G}(x)]
    +w_s(c_x)\\
    &+w_{\mathrm{ttc}}
    \left[1-\frac{\operatorname{TTC}(x)}{\tau_{\mathrm{ttc}}}\right]_+
    +w_q
    \left[1-\frac{y^*(x)}{\tau_q}\right]_+ ,
\end{split}
    \label{eq:memory_utility}
\end{equation}
where $\mathcal{G}(x)$ records whether the sample is mined from Failure (F), Risk (R), or Conflict (C) evidence; $c_x$ is its near-log, recoverable, or far-recoverable state class; $\operatorname{TTC}(x)$ is the minimum time to collision; and $y^*(x)=\max_i y_i$ is the best recovered action score from Eq.~\eqref{eq:r2_score}.

Because the scales of $h_r$ and $u$ vary across rounds, both are min--max normalized over the same candidate pool $\mathcal{P}_r$. Denoting this pool-wise normalization by $\mathcal{N}_{\mathcal{P}_r}(\cdot)$, the memory priority $\rho_r(x)$ is described as: 
\begin{equation}
    \rho_r(x) =
    \alpha\,\mathcal{N}_{\mathcal{P}_r}\!\left(h_r(x)\right)
    +
    \beta\,\mathcal{N}_{\mathcal{P}_r}\!\left(u(x)\right).
    \label{eq:memory_score}
\end{equation}

A global top-$C$ selection rule may cause frequent scenarios or dominant failure types to occupy most of the memory. We therefore adopt a bucket-balanced memory update. Specifically, the candidate pool $\mathcal{P}_r=\operatorname{Unique}(\mathcal{M}_r\cup\mathcal{D}_r)$ is partitioned according to the scene category, the state class $c_x$, and the primary Failure/Risk/Conflict reason. Within each bucket, samples are ranked by the memory priority $\rho_r(x)$, which combines learning difficulty and rollout utility. The limited memory budget is then allocated approximately uniformly across active buckets, and the highest-priority samples in each bucket are retained. If the candidate pool does not exceed the memory capacity, all candidates are kept. This stratified update preserves diverse corrective knowledge across scenarios and corrective knowledge types while still favoring high-priority old and newly retrieved R$^2$ samples. The resulting memory $\mathcal{M}_{r+1}$ serves as the replay set for the next ROCL round.

\begin{algorithm}[!t]
\caption{\method: Rollout-Retrieval Lifelong Policy Learning}
\label{alg:r2lpl}
\begin{algorithmic}[1]
\STATE \textbf{Input:} pretrained planner $\pi_0$, scenarios $\mathcal{S}$, anchor library $\anchors$, memory capacity $C$
\STATE Initialize replay memory $\memory_0 \leftarrow \emptyset$
\FOR{round $r = 0,1,\ldots,R-1$}
    \STATE Roll out $\pi_r$ on scenarios $\mathcal{S}$
    \STATE Mine $\mathcal{I}_r$ from Failure, Risk, and Conflict evidence; stop mining after failure boundaries
    \STATE Initialize recovered dataset $\dataset_r \leftarrow \emptyset$
    \FOR{frame $t \in \mathcal{I}_r$}
        \STATE Restore scene at rollout ego state $\state_t^\pi$
        \STATE Construct admissible candidate anchors and score them by Eq.~\eqref{eq:r2_score}
        \IF{no valid candidate exists}
            \STATE discard frame as unrecoverable
        \ELSE
            \STATE retrieve sparse target $\mathcal{Y}_t$ by Eq.~\eqref{eq:r2_target}
            \STATE add $(\state_t^\pi,\mathcal{Y}_t)$ to $\dataset_r$
        \ENDIF
    \ENDFOR
    \STATE Update $\pi_r$ with LPL objective in Eq.~\eqref{eq:lpl_instantiated}
    \STATE Obtain next planner $\pi_{r+1}$
    \STATE Form $\mathcal{P}_r=\operatorname{Unique}(\memory_r\cup\dataset_r)$ and compute $\rho_r$ by Eqs.~\eqref{eq:memory_difficulty}--\eqref{eq:memory_score}
    \STATE Select $\memory_{r+1}$ by the bucket-balanced rule
\ENDFOR
\STATE \textbf{Return:} lifelong-improved planner $\pi_R$
\end{algorithmic}
\end{algorithm}

\section{Experiments}

\subsection{Benchmarks and Metrics}

We evaluate \method\ on the nuPlan closed-loop simulation benchmarks~\cite{Karnchanachari2024NuplanBenchmark}. NuPlan provides real-world driving logs with rich scenario annotations, long-horizon closed-loop simulation, and standardized planning metrics. These properties allow us to expose policy-induced mistakes through rollout, and evaluate the capability of proposed 
\method\ for continual policy improvement in large-scale real traffic scenarios.

We use three commonly adopted nuPlan benchmarks: \texttt{Val14}, \texttt{Test14-hard}, and \texttt{Test14-random}. \texttt{Val14} contains 1118 validation scenarios from 14 scenario types with an imbalanced category distribution. \texttt{Test14-hard} contains 272 challenging and long-tailed scenarios, while \texttt{Test14-random} contains 261 randomly sampled scenarios with a more balanced distribution over the same 14 scenario types. We consider two closed-loop protocols: non-reactive (NR) and reactive (R) simulation. In NR simulation, the ego vehicle is controlled by the planner while other agents follow logged trajectories. In R simulation, surrounding agents are controlled by an IDM-based reactive model. For each benchmark, \rocl\ data generation and policy updating are performed on the scenarios from that benchmark under the NR protocol. The R results are obtained by directly evaluating the NR-updated checkpoint under the R protocol, without any additional R-specific \rocl\ update. Planner performance is measured by the official nuPlan aggregate score, which combines safety, comfort, and progress-related metrics.

\subsection{Implementation Details}
We instantiate \method{}-base with an anchor-based  planner with transformer encoder-decoder architecture. The planner encodes vectorized ego history, neighboring agents, road geometry, and route context, and predicts scores over a fixed library of $K=4096$ trajectory anchors. Each anchor covers a $4.0$ s planning horizon at $0.2$ s intervals. The model uses a hidden dimension of $256$, an encoder depth of $4$, and a planning decoder depth of $8$. We pretrain the base model on 1M samples from the nuPlan training set for $30$ epochs, using a batch size of $72$, a learning rate of $10^{-4}$, $1000$ warm-up steps, $\lambda_{\mathrm{KL}}=0.1$, and \texttt{bf16-mixed} precision on 4 NVIDIA RTX 4090 GPUs.

Rollouts are performed with model-only planning without any post-processing. Rollout and retrieval are run on two AMD EPYC 7763 64-Core processors. Candidate mining uses a rollout time step of $0.1$ s. Failure-related frames are collected from the $40$ steps preceding the failure frame. A failure frame is identified when collision or out-of-road is detected. Risk frames are identified by a minimum TTC below $1.0$ s, together with a $10$-step pre-risk window. Conflict frames are selected according to misalignment between  policy-induced states and logged expert states, using thresholds of $0.5$ m/s for waiting speed, $3.0$ m for lag-progress gap, and $1.0$ m/s for moving speed.

For R$^2$ retrieval, we first prefilter at most $1024$ geometry-compatible anchors before scoring. Compatibility is determined by a maximum reference-path distance of $6.0$ m and a maximum heading error of $60^\circ$. A frame is discarded as unrecoverable if no valid anchor is found. Otherwise, the top $256$ recovered anchors are stored as sparse replay targets. State classes are defined by $\delta_{\rm near}^{xy}=1.0$ m, $\delta_{\rm near}^{\psi}=15^\circ$, $\delta_{\rm near}^{v}=2.0$ m/s, $\delta_{\rm rec}^{xy}=5.0$ m, and $\delta_{\rm rec}^{\psi}=45^\circ$. Near-log retrieval further requires the expert-consistency score to be within $0.05$ of the best score and at least $0.75$. The final anchor score in Eq.~\eqref{eq:r2_score} uses $(w_{\rm sim}, w_{\rm ref}, w_{\rm exp})=(0.10,0.10,0.80)$ for near-log states, $(0.65,0.30,0.05)$ for recoverable states, and $(0.80,0.20,0.00)$ for far-off-policy states.

For LPL, we use DER with scenario-aware rollout-priority replay as the default continual learning method. Each \rocl\ round is trained for $20$ epochs with a batch size $24$ and a memory capacity $C=4096$. In Eq.~\eqref{eq:lpl_instantiated}, the replay loss weight is $\lambda_{\rm rep}=1.0$, the DER distillation weight is $\lambda_{\rm DER}=0.25$, and the distillation temperature is $1.0$. In Eq.~\eqref{eq:memory_score}, the normalized training-loss component and the rollout-utility component are both assigned a weight of $1.0$. Expert-cache mixing uses a ratio of $0.25$, a loss weight of $\lambda_E=0.25$, and at most $20{,}000$ expert samples per task.

\subsection{Comparison with Baselines}

We compare \method\ with representative learning-based planners on the nuPlan benchmarks to evaluate the relative performance improvement and to assess its competitiveness among existing policy learning paradigms. The comparison focuses on planners whose reported performance mainly comes from learned policy modeling, including recent generative methods and reinforcement-learning fine-tuning methods. Since this experiment aims to compare learned policy capability rather than rule-assisted execution quality, we do not include methods whose final scores rely heavily on rule-based post-processing, trajectory repair, or hybrid rule-based refinement. Such methods are important for engineering robustness, but their performance may reflect both learned policy quality and external correction modules, making it difficult to isolate the capability of the learned planner. The compared methods include:

\noindent$\bullet$ \textbf{UrbanDriver}~\cite{Scheel2021UrbanDriver} is an early learning-based planner that optimizes a driving policy from real-world demonstrations using a policy-gradient objective.\\
\noindent$\bullet$ \textbf{PDM-Open}~\cite{Dauner2023PartingWithMisconceptions} predicts future way points using an IDM-based centerline and ego history.\\
\noindent$\bullet$ \textbf{PlanTF}~\cite{cheng2023plantf} is an imitation-based transformer planner trained to predict expert future motion.\\
\noindent$\bullet$ \textbf{PLUTO}~\cite{cheng2024pluto} improves imitation-based planning with stronger scene representation, data augmentation, and training design.\\
\noindent$\bullet$ \textbf{Diffusion Planner}~\cite{diffusionplanner2025zheng} formulates planning as conditional diffusion trajectory generation with guidance.\\
\noindent$\bullet$ \textbf{Flow Planner}~\cite{tan2025flow} applies flow matching and interaction-aware modeling for tokenized trajectory generation.\\
\noindent$\bullet$ \textbf{DFP}~\cite{Zhang2026DFP} uses diffusion forcing with history-annealed planning, and \textbf{DFP-FM} augments it with flow matching.\\
\noindent$\bullet$ \textbf{Plan-R1}~\cite{tang2026planr1} casts trajectory planning as motion-token language modeling and applies rule-based GRPO post-training.

We report our pretrained anchor-based planner as \method-base, a fixed-budget five-round version as \method-\rocl-5, and the best observed result within the ten-round search budget as \method{}-ROCL-10-best.

\begin{table*}[!t]
\caption{Main closed-loop results on nuPlan benchmarks. NR and R denote non-reactive and reactive simulation protocols, respectively. Scores are aggregated metrics, where higher values indicate better performance. \textbf{Bold} denote the best and \underline{underlined} the second-best results among compared methods.}
\label{tab:main_results}
\centering
\footnotesize
\setlength{\tabcolsep}{4.2pt}
\renewcommand{\arraystretch}{1.05}
\begin{tabular*}{\linewidth}{@{\extracolsep{\fill}}llcccccc@{}}
\toprule
\multirow{2}{*}[-0.45ex]{\textbf{Method}} & \multirow{2}{*}[-0.45ex]{\textbf{Training Paradigm}} &
\multicolumn{2}{c}{\textbf{Val14}} &
\multicolumn{2}{c}{\textbf{Test14-hard}} &
\multicolumn{2}{c}{\textbf{Test14-random}} \\
\cmidrule(lr){3-4}\cmidrule(lr){5-6}\cmidrule(lr){7-8}
& & \textbf{NR} & \textbf{R} & \textbf{NR} & \textbf{R} & \textbf{NR} & \textbf{R} \\
\midrule
\textcolor{gray}{Expert} & \textcolor{gray}{Human log replay} & \textcolor{gray}{93.53} & \textcolor{gray}{80.32} & \textcolor{gray}{85.96} & \textcolor{gray}{68.80} & \textcolor{gray}{94.03} & \textcolor{gray}{75.86} \\
\midrule
UrbanDriver~\cite{Scheel2021UrbanDriver} & Imitation learning & 68.57 &  64.11 & 50.40 & 49.95 & 51.83 & 67.15 \\
PDM-Open~\cite{Dauner2023PartingWithMisconceptions} & Imitation learning & 53.53 & 54.24 & 33.51 & 35.83 & 52.81 & 57.23 \\
PlanTF~\cite{cheng2023plantf} & Imitation learning & 84.27 & 76.95 & 69.70 & 61.61 & 85.62 & 79.58 \\
PLUTO~\cite{cheng2024pluto} & Imitation learning & 88.89 & 78.11 & 70.03 & 59.74 & 89.90 & 78.62 \\
Diffusion Planner~\cite{diffusionplanner2025zheng} & Generative IL & 89.87 & 82.80 & 75.99 & 69.22 & 89.19 & 82.93 \\
Flow Planner~\cite{tan2025flow} & Generative IL & 90.43 & 83.31 & 76.47 & 70.42 & 89.88 & 82.93 \\
DFP~\cite{Zhang2026DFP} & Generative IL & 90.33 & 79.97 & 76.91 & 63.56 & 90.69 & 81.96 \\
DFP-FM~\cite{Zhang2026DFP} & Generative IL & \textbf{92.68} & 81.30 & \underline{79.43} & 67.94 & 90.62 & 83.59 \\
Plan-R1~\cite{tang2026planr1} & IL + RL alignment & 88.98 & \textbf{87.69} & 77.45 & \underline{77.20} & \underline{91.23} & \textbf{90.04} \\
\method{}-base (ours) & Imitation learning  & 75.39 & 73.87 & 60.67 & 65.25 & 70.74 & 72.96 \\
\method{}-ROCL-5 (ours) & IL + R²LPL & \underline{91.26} & \underline{85.38} & \textbf{83.51} & \textbf{78.38} & \textbf{92.25} & \underline{87.99} \\
\midrule
\method{}-ROCL-10-best & IL + R²LPL (envelope) & 92.22 & 85.83 & 86.54 & 78.88 & 93.94 & 88.20 \\
\bottomrule
\end{tabular*}
\end{table*}

Table~\ref{tab:main_results} reports the main closed-loop results on all three nuPlan benchmarks. The pretrained anchor-based planner is relatively weak compared with recent learning-based planners, achieving only 75.39/73.87, 60.67/65.25, and 70.74/72.96 on \texttt{Val14}, \texttt{Test14-hard}, and \texttt{Test14-random} under NR/R protocols. After five \rocl\ rounds, the same planner is improved to 91.26/85.38, 83.51/78.38, and 92.25/87.99, yielding absolute gains of $+15.87$, $+22.84$, and $+21.51$ points under NR and $+11.51$, $+13.13$, and $+15.03$ points under R. These improvements elevate a moderate base planner to the first or second place among all compared learning-based methods on every benchmark and protocol. Since \rocl\ updates the policy without changing the base architecture or adding deployment-time rule refinement, the results indicate that the proposed framework improves the learned policy itself rather than relying on a stronger planner design or external execution-time correction.

Among the three splits, \texttt{Test14-hard} provides the clearest stress test for closed-loop policy improvement because it focuses on challenging and long-tailed scenarios. On this split, \method-\rocl-5 achieves 83.51 NR and 78.38 R, outperforming the strongest listed learning-based baselines by $4.08$ and $1.18$ points, respectively. The larger margin under NR is expected, since both \rocl\ data generation and policy updating are conducted with NR rollouts. Nevertheless, the R score also increases from 65.25 to 78.38 without any R-specific \rocl\ update, reaching the best result among all compared methods. This cross-protocol improvement suggests that the retrieved corrective knowledge is not merely tied to the fixed-agent trajectories used during NR rollout. Since \rocl\ supervises recoverable mistake-related states rather than directly imitating a specific closed-loop trajectory, the learned corrections can encode more general decision preferences for avoiding unsafe or conflicting behaviors, which remain beneficial even when the interaction dynamics changes.

The results on \texttt{Val14} and \texttt{Test14-random} show a more moderate but consistent pattern. These splits cover broader evaluation scenarios rather than explicitly concentrating on hard long-tail cases, so the advantage of mistake-driven correction is less pronounced than on \texttt{Test14-hard}. Nevertheless, \method-\rocl-5 still ranks second on \texttt{Val14} under both protocols and first/second on \texttt{Test14-random} under NR/R. This indicates that correcting policy-induced mistakes does not sacrifice broader closed-loop competence. In other words, \method\ is not only effective on failure-prone scenarios, but also retains strong general performance after lifelong updates.

The learning-envelope result \method{}-ROCL-10-best further shows that additional improvement is possible within the same ten-round search budget, reaching 92.22/85.83, 86.54/78.88, and 93.94/88.20 on the three splits. We use \method-\rocl-5 as the primary fixed-budget result and report \method{}-ROCL-10-best only to indicate the potential performance envelope of the iterative improvement process.

\begin{table*}[!t]
\caption{Detailed \texttt{Test14-hard} closed-loop metrics across five \rocl\ updates. Scores are aggregated metrics, where higher values indicate better performance. Parenthesized values report absolute changes from the previous round.}
\label{tab:rocl_round_details}
\centering
\footnotesize
\setlength{\tabcolsep}{4.0pt}
\renewcommand{\arraystretch}{1.05}
\newcommand{\basegain}{\scriptsize\textcolor{gray}{\phantom{(+00.00)}}}
\newcommand{\gain}[1]{\scriptsize\textcolor{gray}{(#1)}}
\begin{tabular}{@{}cllllll@{}}
\toprule
\textbf{\rocl\ Round} & \textbf{Score}  & \textbf{Collisions} & \textbf{TTC} & \textbf{Drivable} & \textbf{Comfort} & \textbf{Progress} \\
\midrule
0 (Base) & 60.67 \basegain & 75.55 \basegain & 64.76 \basegain & 90.81 \basegain & 88.24 \basegain & 94.81 \basegain \\

1 & 71.64 \gain{+10.97} & 91.36 \gain{+15.81} & 84.92 \gain{+20.16} & 95.95 \gain{+5.14} & 90.81 \gain{+2.57} & 74.68 \gain{-20.13} \\

2 & 78.63 \gain{+6.99} & 91.54 \gain{+0.18} & 82.72 \gain{-2.20} & 95.59 \gain{-0.36} & 89.34 \gain{-1.47} & 87.78 \gain{+13.1} \\

3 & 80.70 \gain{+2.07} & 94.85 \gain{+3.31} & 84.56 \gain{+1.84} & 95.22 \gain{-0.37} & 90.44 \gain{+1.10} & 88.87 \gain{+1.09} \\

4 & 83.00 \gain{+2.30} & 94.49 \gain{-0.36} & 85.66 \gain{+1.10} & 97.43 \gain{+2.21} & 92.28 \gain{+1.84} & 90.16 \gain{+1.29} \\

5 & 83.52 \gain{+0.52} & 93.20 \gain{-1.29} & 85.29 \gain{-0.37} & 97.43 \gain{+0.00} & 91.54 \gain{-0.74} & 91.80 \gain{+1.64} \\
\bottomrule
\end{tabular}
\end{table*}

\begin{figure}[!t]
\centering
\includegraphics[width=\columnwidth]{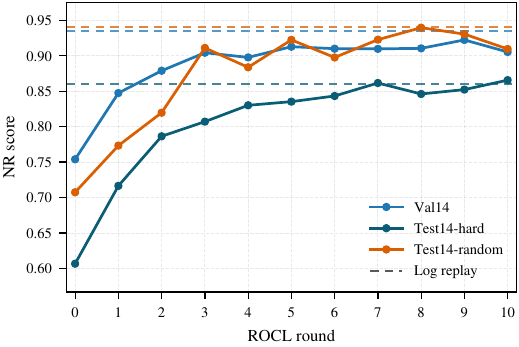}
\caption{Closed-loop performance over \rocl\ rounds on \texttt{Val14}, \texttt{Test14-hard}, and \texttt{Test14-random}. Solid curves report \method\ scores across \rocl\ rounds, and dashed horizontal lines report the corresponding expert-log replay scores.}
\label{fig:iter_curve}
\end{figure}

\begin{figure*}[!t]
\centering
\includegraphics[width=\textwidth]{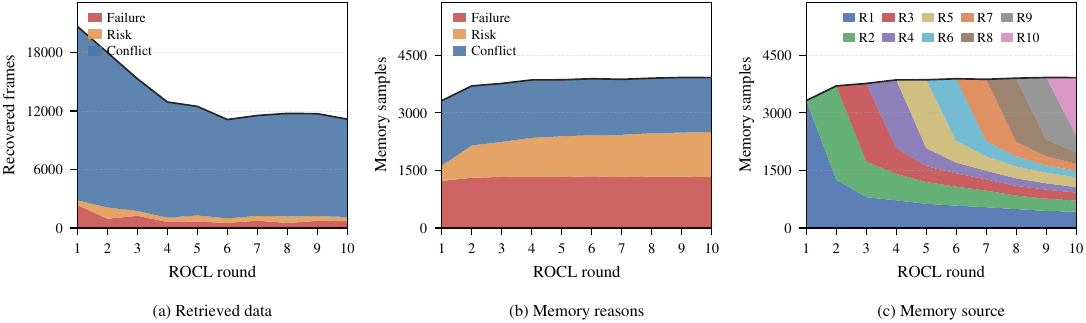}
\caption{Evolution of policy-induced failure data and replay memory across \rocl\ rounds. The three panels summarize: (a)  retrieved data classified by recovery reason, (b) memory composition by recovery reason, and  (c) memory source over each \rocl\ round.}
\label{fig:data_evolution}
\end{figure*}

\subsection{Analysis on Iterative Improvement}
This subsection examines the iterative behavior of \method. The analysis focuses on two empirical aspects: whether the planner continues to improve across repeated rollout-learning rounds, and what data and replay mechanisms account for this improvement. We use the round-wise score curves to address the former, and the mined-data and memory statistics to analyze the latter.

Fig.~\ref{fig:iter_curve} shows that the improvement is not limited to the first update. On \texttt{Test14-hard}, the NR score increases sharply in the early rounds and continues to improve more gradually afterwards, eventually exceeding the expert-log replay reference. Although the curve is not strictly monotonic, such fluctuations are expected because each policy update changes the induced closed-loop state distribution and may shift the balance among safety, comfort, and progress terms in the nuPlan aggregate score. Nevertheless, the overall upward trend indicates that \method\ continues to accumulate useful corrective knowledge across ROCL rounds. The smoother curves on \texttt{Val14} and \texttt{Test14-random} further indicate that iterative rollout learning does not only benefit the hardest subset, although the largest gain appears on \texttt{Test14-hard}, where the initial policy exposes more severe residual errors.

To understand how the improvement is obtained, Table~\ref{tab:rocl_round_details} reports the detailed metric evolution on \texttt{Test14-hard}. Parenthesized values denote absolute changes from the previous round, because the nuPlan aggregate and component metrics are bounded scores.

The metric breakdown suggests a two-stage improvement pattern. The first round gives the largest aggregate gain, from 60.67 to 71.64, mainly through safety-related metrics: collision score improves by 15.81 points and TTC score improves by 20.16 points. This is consistent with the role of rollout mining, since the base policy exposes safety-critical states that can be converted into corrective targets when they are recoverable. Subsequent rounds provide smaller but still meaningful gains, with improvements distributed over drivable-area compliance, comfort, and progress. The temporary progress drop after the first round also illustrates the multi-objective nature of closed-loop planning: an update that strongly suppresses risky behavior can initially make the policy more conservative, and later rounds restore the progress score while preserving the safety gains. By \rocl\ round 5, the aggregate score reaches 83.52, with relatively high collision, TTC, drivable-area, comfort, and progress scores retained simultaneously.

The remaining question is why repeated updates continue to provide useful supervision. Fig.~\ref{fig:data_evolution} reports three statistics related to the data stream and memory: the retrieved training data mined in each round, the composition of the replay memory, and the source rounds preserved by the memory. These statistics indicate whether later policies still expose informative errors and whether the learner maintains access to previously recovered corrections.

Fig.~\ref{fig:data_evolution}(a) shows that the recovered data are policy-dependent rather than fixed. \emph{Failure} denotes frames in the temporal window before an actual rollout failure, \emph{Risk} denotes high-risk context frames such as low TTC or dangerous near-miss states, and \emph{Conflict} denotes model-expert disagreement frames where the policy behavior differs substantially from the logged expert. Early rounds contain many failure-window corrections, while later rounds still provide risk and conflict samples around the remaining difficult states. Thus, after the most obvious failures are corrected, rollout continues to reveal residual mistakes that can be used for further learning.

Fig.~\ref{fig:data_evolution}(b) and Fig.~\ref{fig:data_evolution}(c) show how LPL stabilizes this evolving data stream. The memory keeps a mixed composition of Failure, Risk, and Conflict samples rather than being dominated by one category, and it preserves samples from multiple source rounds. Later updates are therefore not simple fine-tuning on the newest rollout cache. They combine newly exposed residual mistakes with replayed corrections from previous policies. This interaction between policy-dependent mining and replay explains the sustained improvement observed in Fig.~\ref{fig:iter_curve}: R$^2$ supplies newly exploited recoverable mistakes as the policy updates, while LPL prevents earlier experience from being overwritten.

\subsection{Ablation Studies}
We organize the ablation around two questions. The first is whether the improvement is attributed to overfitting target-scenario expert knowledge rather than learning from policy-induced mistakes. The second is whether recoverability-aware supervised fine-tuning is sufficient, or whether lifelong policy learning is needed for stable multi-round continuous improvement.

\begin{table}[!t]
\caption{Ablation on \texttt{Test14-hard} under a five-round \rocl\ budget. SFT denotes supervised fine-tuning, and w/o denotes without. Round 0 is the same pretrained base policy for all variants. The target expert-log SFT baseline is a one-shot adaptation baseline, so later \rocl\ rounds are not applicable.}
\label{tab:ablation}
\centering
\footnotesize
\setlength{\tabcolsep}{0pt}
\renewcommand{\arraystretch}{1.08}
\begin{tabular*}{\columnwidth}{@{\extracolsep{\fill}}lcccccc@{}}
\toprule
\multirow{2}{*}{\textbf{Method}} & \multicolumn{6}{c}{\textbf{\rocl\ Score}} \\
\cmidrule(lr){2-7}
& \textbf{0} & \textbf{1} & \textbf{2} & \textbf{3} & \textbf{4} & \textbf{5} \\
\midrule
Base + expert-log SFT & 60.67 & 60.79 & -- & -- & -- & -- \\
Base + RoaD-style SFT & 60.67 & 66.36 & 67.34 & 66.62 & 65.57 & 63.58 \\
Base + R$^2$-style SFT & 60.67 & 71.29 & 75.37 & 76.12 & 79.51 & 76.17 \\
\method\ w/o expert mix & 60.67 & 72.44 & 78.34 & 80.89 & 83.18 & 82.64 \\
\method\ full & 60.67 & 71.64 & 78.63 & 80.70 & 83.00 & 83.52 \\
\bottomrule
\end{tabular*}
\end{table}

Table~\ref{tab:ablation} separates the effects of target-scenario expert knowledge, closed-loop rollout data, recoverability-aware retrieval, and lifelong policy learning. Directly fine-tuning the base model on target-scenario expert logs provides almost no improvement, increasing the score from 60.67 to 60.79. This suggests that the main gain in \method\ is not simply due to additional exposure to the target scenarios or memorization of their logged expert behavior.

We further compare with a RoaD-style SFT baseline. This baseline uses closed-loop rollout states and the corresponding expert demonstration guided targets as supervised training samples, but does not perform R$^2$ recoverability-aware target retrieval or LPL. It improves over expert-log SFT and reaches 67.34 after two rounds, showing that policy-induced states and targets recovered from expert logs do contain useful information. However, its gain is limited. A likely reason is the uneven quality of the induced supervision. When the policy has deviated substantially from the logged expert state, forcing the rollout state toward an expert-aligned demonstration can introduce misleading targets, because the logged expert motion may no longer be a feasible or appropriate correction from the learner-induced state. The later score decrease to 63.58 then reflects a stability and forgetting issue under repeated updates on such noisy rollout-derived targets.

Replacing RoaD-style targets with R$^2$ retrieval provides a stronger SFT baseline. Base model trained with R$^2$-style SFT reaches 71.29 after the first round, close to the 71.64 obtained by full \method, which indicates that recoverability-aware target construction provides the main initial corrective signal. However, this SFT baseline improves more slowly in later rounds and drops from 79.51 at round 4 to 76.17 at round 5. This behavior is consistent with learning under insufficient retention: while new R$^2$ knowledge is being fitted, earlier correction knowledge are not explicitly preserved, and later updates cause the polity to forget more than it learns. In contrast, \method\ maintains scores above 80 after round 3 and reaches 83.52 at round 5, showing the role of LPL in retaining previously acquired correction knowledge while incorporating new R$^2$ knowledge. The variant without expert mix is close to the full model and is slightly higher in some intermediate rounds, indicating that expert-log mixing is not the primary source of the closed-loop performance gain. Overall, the ablation supports that \method\ benefits from both high-quality recoverable supervision from R$^2$ and non-forgetting policy update through LPL, while expert mixing mainly acts as a conservative regularizer.

\subsection{Qualitative Analysis}
We visualize representative base-policy failures and the corresponding round-5 behaviors after \method\ updates. The  qualitative analysis is shown in three aspects: how R$^2$ corrective knowledge is retrieved, which rollout states are used by R$^2$ for recovery knowledge construction, and to illustrate how the improved policy avoids repeating the same failure.

\begin{figure}[!t]
\centering
\subfloat[High lateral acceleration]{
    \includegraphics[width=0.46\textwidth]{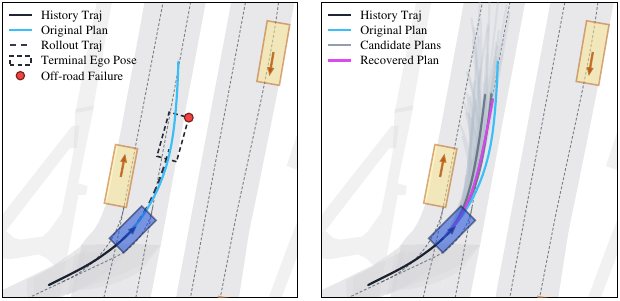}
}
\vspace{0.35em}

\subfloat[Changing lane]{
    \includegraphics[width=0.46\textwidth]{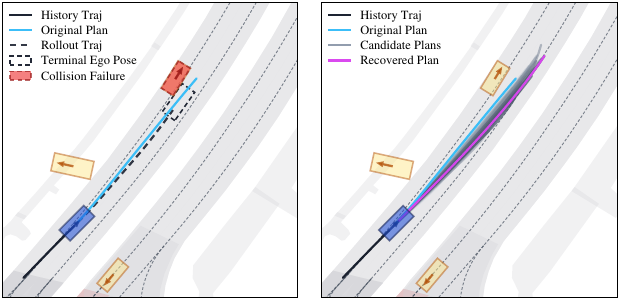}
}
\vspace{0.35em}

\subfloat[Near pedestrian on crosswalk]{
    \includegraphics[width=0.46\textwidth]{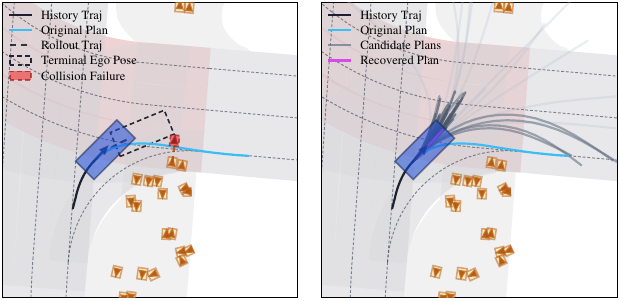}
}
\caption{Qualitative examples of R$^2$ corrective-supervision construction, \textcolor{blue}{Blue} boxes denote the rollout ego vehicle, and \textcolor{orange}{yellow} boxes denote surrounding agents. Case (a) shows a high-lateral-acceleration maneuver that eventually leads to an off-road failure, case (b) shows a lane-changing collision failure, and case (c) shows a collision failure with pedestrians. Each row shows a recoverable state mined from a base-policy failure and the corresponding retrieved target in the same state.}
\label{fig:r2_retrieval_examples}
\end{figure}

Fig.~\ref{fig:r2_retrieval_examples} provides a qualitative visualization of how R$^2$ constructs corrective supervision from closed-loop failures. For each failure case, R$^2$ first mines recoverable policy-induced states before the terminal failure. The left panels show the base-policy behavior around one of these states: although the original plan predicted at the selected frame may appear locally plausible, the subsequent closed-loop rollout follows this decision and eventually leads to an off-road or collision failure. Therefore, R$^2$ focuses on the pre-failure window to identify recoverable states and correct the decisions that drive the policy toward failure, rather than simply labeling the terminal failure pose. Given the recoverable state, R$^2$ then searches the structured action space for alternative corrective plans. As shown in the right panels, candidate plans are evaluated by the normalized target score $y_i$ in Eq.~\eqref{eq}, with darker candidates indicating higher-scoring plans under the evaluator. The selected recovered plan provides a safer and more feasible alternative to the original plan, and is therefore used as the retrieved target. This target is stored as R$^2$ knowledge and used as corrective supervision in the subsequent lifelong policy update.

\begin{figure*}[!t]
\centering
\subfloat[High lateral acceleration]{
    \includegraphics[width=0.98\linewidth]{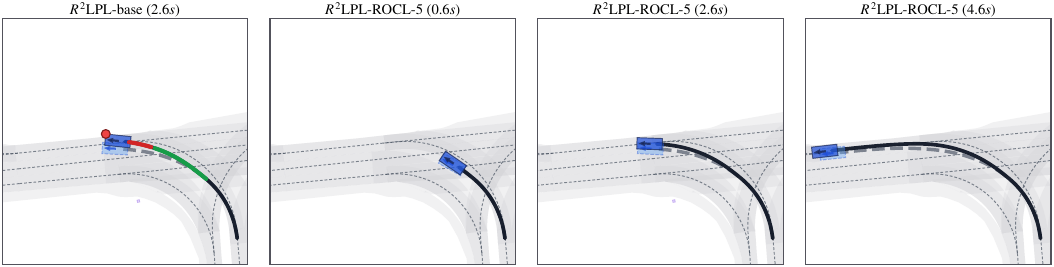}
}
\vspace{0.35em}

\subfloat[Low magnitude speed]{
    \includegraphics[width=0.98\linewidth]{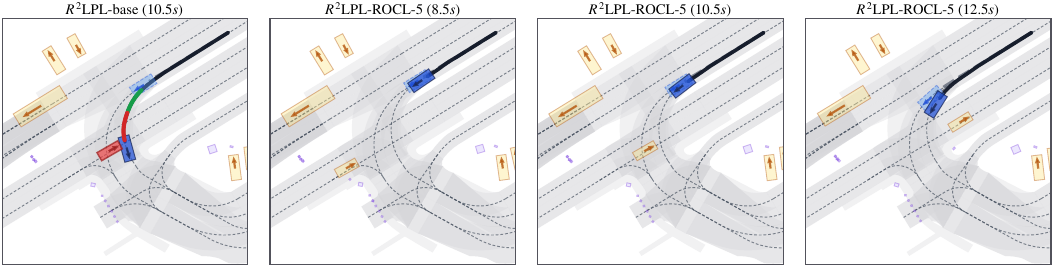}
}
\vspace{0.35em}

\subfloat[Traversing pickup-dropoff]{
    \includegraphics[width=0.98\linewidth]{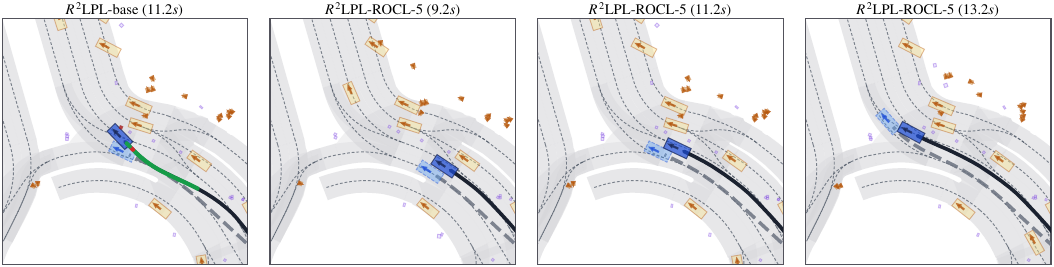}
}
\caption{Qualitative recovery examples. Each row shows a base-policy failure and the corresponding \rocl\ round-5 \method\ rollout results around the same critical frame. Case (a) corresponds to an off-road failure associated with high lateral acceleration, while cases (b) and (c) correspond to collision failures. \textcolor{blue}{Blue} boxes denote the rollout ego vehicle, \textcolor{cyan}{light-blue} dashed boxes denote the log expert, \textcolor{orange}{yellow} boxes denote surrounding agents, and \textcolor{purple}{purple} boxes denote static obstacles. Black solid and gray dashed curves show rollout and expert histories, respectively. \textcolor{green!60!black}{Green} and \textcolor{red}{red} thick trajectory segments in the base rollout indicate recoverable and unrecoverable failure-window states. The round-5 panels at $-2$ s, $0$ s, and $+2$ s show how the improved policy behaves before, at, and after the critical frame.}
\label{fig:qualitative}
\end{figure*}

Fig.~\ref{fig:qualitative} provides concrete examples of the rollout-retrieval process. In the base-policy rollout, R$^2$ does not treat all states before failure equally. Green segments indicate states for which the action set and evaluator can still identify feasible recovery targets, and these states are used to construct R$^2$ knowledge. Red segments indicate states that have become unrecoverable under the available action space or scene constraints, and are therefore excluded from training. This visualization corresponds directly to the recoverable/unrecoverable separation in the method. It also illustrates why \method\ mines a temporal window before failure rather than only the final failure frame. In closed-loop planning, a collision or off-road event is often the consequence of erroneous decisions made several seconds earlier. When close to the final failure frame, the failure may already be irreversible. Corrective learning therefore needs to identify and repair the preceding recoverable decisions that lead the vehicle toward the dangerous state.

The three cases in Fig.~\ref{fig:qualitative} cover different failure modes. In Fig.~\ref{fig:qualitative}(a), the base policy drives the ego vehicle outside the drivable region while exhibiting high lateral acceleration. In Fig.~\ref{fig:qualitative}(b) and Fig.~\ref{fig:qualitative}(c), the base policy leads to collision-prone interactions, including collision with other agent in left turn and collision with a static obstacle. The round-5 panels show the learned policy at $-2$s, $0$s, and $+2$s relative to the base policy's failing moment. Compared with the base policy, the improved policy adjusts its behavior before reaching the failure state and no longer produces the same off-road or collision outcome. These examples illustrate how R$^2$ supplies policy-specified recovery knowledge from selected failure-window states, and how LPL turns this knowledge into corrected closed-loop behavior in later rollouts.

\section{Conclusion}

We presented \method, a rollout-retrieval lifelong policy learning paradigm for autonomous driving. Instead of treating closed-loop failures only as evaluation outcomes, \method\ uses policy-induced rollouts to identify mistake-related states, retrieve recoverable corrective targets from a structured action space, and incorporate the retrieved knowledge through lifelong policy learning. The central insight is that the most useful knowledge for continual policy improvement is not the failure itself, but the corrective knowledge that helps the policy avoid repeating the same mistakes. Experiments on nuPlan show that \method\ can substantially improve an anchor-based planner with moderate performance and elevate it to state-of-the-art performance among learning-based planners across multiple closed-loop benchmarks. The gains are most pronounced on hard and long-tailed scenarios, while the updated policy also maintains strong performance on broader validation and random test splits. Moreover, although the corrective knowledge is acquired from non-reactive rollouts, the improved policy generalizes to reactive simulation, suggesting that the retrieved targets encode reusable decision preferences rather than overfitting to non-reactive agent patterns. These results demonstrate that large-scale closed-loop policy improvement can be achieved with compact \rocl\ rounds through \method, without relying solely on larger expert-log imitation, reward-driven exploration, or deployment-time rule refinement. 

The current instantiation of \method\ focuses on decision policies with explicit or sampleable action candidates, where recoverability-aware target retrieval can be implemented through candidate evaluation. Its effectiveness therefore depends on the coverage of the action space and the alignment of the evaluator used to select corrective targets. Extending this idea to continuous generative planners may require retrieval or projection in latent trajectory space, together with stronger mechanisms for filtering noisy targets. Another important direction is to replace or complement simulation-based rollouts with learned physical world models with human-like agents, enabling policies to acquire corrective knowledge from a much larger and more diverse stream of closed-loop interactions. 

\section*{Acknowledgments}
Large language models, including models from DeepSeek and GitHub Copilot, were used for code debugging, language polishing, and readability improvement. The authors verified and take full responsibility for all presented contents, experimental results, and conclusions. All figures and corresponding experimental results can be reproduced using the code and instructions available in our public repository: \url{https://github.com/Engibacter/R2LPL}.

\bibliographystyle{IEEEtran}
\bibliography{R2LPL_refs}

@inproceedings{Hu2023UniAD,
  title={Planning-oriented Autonomous Driving},
  author={Hu, Yihan and Yang, Jiazhi and Chen, Li and Li, Keyu and Sima, Chonghao and Zhu, Xizhou and Chai, Siqi and Du, Senyao and Lin, Tianwei and Wang, Wenhai and Lu, Lewei and Jia, Xiaosong and Liu, Qiang and Dai, Jifeng and Qiao, Yu and Li, Hongyang},
  booktitle={Proceedings of the IEEE/CVF Conference on Computer Vision and Pattern Recognition },
  pages={17853--17862},
  year={2023}
}

@article{Chitta2023Transfuser,
   author = {Chitta, K. and Prakash, A. and Jaeger, B. and Yu, Z. and Renz, K. and Geiger, A.},
   title = {TransFuser: Imitation With Transformer-Based Sensor Fusion for Autonomous Driving},
   journal = {IEEE Transactions on Pattern Analysis and Machine Intelligence},
   volume = {45},
   number = {11},
   pages = {12878--12895},
   DOI = {10.1109/TPAMI.2022.3200245},
   year = {2023},
}

@article{e2esurvey2024chen,
  author={Chen, Li and Wu, Penghao and Chitta, Kashyap and Jaeger, Bernhard and Geiger, Andreas and Li, Hongyang},
  journal={IEEE Transactions on Pattern Analysis and Machine Intelligence}, 
  title={End-to-End Autonomous Driving: Challenges and Frontiers}, 
  year={2024},
  volume={46},
  number={12},
  pages={10164-10183},
  doi={10.1109/TPAMI.2024.3435937}
}

@inproceedings{Dauner2024NAVSIM,
	title = {NAVSIM: Data-Driven Non-Reactive Autonomous Vehicle Simulation and Benchmarking},
	author = {Daniel Dauner and Marcel Hallgarten and Tianyu Li and Xinshuo Weng and Zhiyu Huang and Zetong Yang and Hongyang Li and Igor Gilitschenski and Boris Ivanovic and Marco Pavone and Andreas Geiger and Kashyap Chitta},
	booktitle = {Advances in Neural Information Processing Systems},
	year = {2024},
    pages = {28706--28719},
    volume = {37},
}

@inproceedings{sun2025sparsedrive,
  title={Sparsedrive: End-to-end autonomous driving via sparse scene representation},
  author={Sun, Wenchao and Lin, Xuewu and Shi, Yining and Zhang, Chuang and Wu, Haoran and Zheng, Sifa},
  booktitle={Proceedings of the IEEE International Conference on Robotics and Automation},
  pages={8795--8801},
  year={2025},
}

@article{DenseLearning2026,
  author  = {Feng, Shuo and Zhu, Haojie and Sun, Haowei and Yan, Xintao and He, Linxuan and Yang, Jingxuan and Su, Guangzhen and Li, Boqi and Li, Shu and Wang, Ling and Shen, Shengyin and Liu, Henry X.},
  title   = {Breaking through Safety Performance Stagnation in Autonomous Vehicles with Dense Learning},
  journal = {Nature Communications},
  volume  = {17},
  number  = {3163},
  year    = {2026}
}

@InProceedings{Scheel2021UrbanDriver,
title = 	 {Urban Driver: Learning to Drive from Real-world Demonstrations Using Policy Gradients},
author =       {Scheel, Oliver and Bergamini, Luca and Wolczyk, Maciej and Osi\'nski, B\l{a}\.{z}ej and Ondruska, Peter},
booktitle = 	 {Proceedings of the Conference on Robot Learning},
pages = 	 {718--728},
year = 	 {2022},
volume = 	 {164},
}

@InProceedings{mp32021casas,
   author = {Casas, Sergio and Sadat, Abbas and Urtasun, Raquel},
   title = {MP3: A Unified Model to Map, Perceive, Predict and Plan},
   booktitle = {Proceedings of the IEEE/CVF Conference on Computer Vision and Pattern Recognition},
   pages = {14398-14407},
   year = {2021},
}

@InProceedings{Huang_2023_ICCV,
    author    = {Huang, Zhiyu and Liu, Haochen and Lv, Chen},
    title     = {GameFormer: Game-theoretic Modeling and Learning of Transformer-based Interactive Prediction and Planning for Autonomous Driving},
    booktitle = {Proceedings of the IEEE/CVF International Conference on Computer Vision},
    year      = {2023},
    pages     = {3903--3913}
}

@InProceedings{Dauner2023PartingWithMisconceptions,
  title = 	 {Parting with Misconceptions about Learning-based Vehicle Motion Planning},
  author =       {Dauner, Daniel and Hallgarten, Marcel and Geiger, Andreas and Chitta, Kashyap},
  booktitle = 	 {Proceedings of The Conference on Robot Learning},
  pages = 	 {1268--1281},
  year = 	 {2023},
  volume = 	{229}
}

@article{cheng2024pluto,
      title={{PLUTO}: Pushing the Limit of Imitation Learning-based Planning for Autonomous Driving}, 
      author={Jie Cheng and Yingbing Chen and Qifeng Chen},
      year={2024},
      journal={arXiv preprint arXiv:2404.14327}
}

@INPROCEEDINGS{cheng2023plantf,
      title={Rethinking Imitation-based Planner for Autonomous Driving},
      author={Jie Cheng and Yingbing Chen and Xiaodong Mei and Bowen Yang and Bo Li and Ming Liu},
      booktitle={Proceedings of the IEEE International Conference on Robotics and Automation}, 
      pages={14123--14130},
      year={2024}
}

@inproceedings{vad2023jiang,
   author = {Jiang, Bo and Chen, Shaoyu and Xu, Qing and Liao, Bencheng and Chen, Jiajie and Zhou, Helong and Zhang, Qian and Liu, Wenyu and Huang, Chang and Wang, Xinggang},
   title = {VAD: Vectorized Scene Representation for Efficient Autonomous Driving},
   booktitle = {Proceedings of the IEEE/CVF International Conference on Computer Vision},
   pages = {8340--8350},
   year = {2023},
}

@article{hydramdp2024li,
  title={Hydra-MDP: End-to-end Multimodal Planning with Multi-target Hydra-Distillation},
  author={Li, Zhenxin and Li, Kailin and Wang, Shihao and Lan, Shiyi and Yu, Zhiding and Ji, Yishen and Li, Zhiqi and Zhu, Ziyue and Kautz, Jan and Wu, Zuxuan and others},
  year={2024},
  journal={arXiv preprint arXiv:2406.06978},
}

@inproceedings{chen2024vadv2,
  title={Vadv2: End-to-end vectorized autonomous driving via probabilistic planning},
  author={Chen, Shaoyu and Jiang, Bo and Gao, Hao and Liao, Bencheng and Xu, Qing and Zhang, Qian and Huang, Chang and Liu, Wenyu and Wang, Xinggang},
  year={2024},
  booktitle={Proceedings of the International Conference on Learning Representations}
}

@inproceedings{Sun2025GeneralizingMotionPlanners,
   author = {Sun, Q. and Wang, H. and Zhan, J. and Nie, F. and Wen, X. and Xu, L. and Zhan, K. and Jia, P. and Lang, X. and Zhao, H.},
   title = {Generalizing Motion Planners with Mixture of Experts for Autonomous Driving},
   booktitle = {Proceedings of the IEEE International Conference on Robotics and Automation},
   year = {2025},
   pages = {6033--6039},
}

@inproceedings{OpenDriveVLA2025,
  title={OpenDriveVLA: Towards End-to-end Autonomous Driving with Large Vision Language Action Model},
  author={Zhou, Xingcheng and Han, Xuyuan and Yang, Feng and Ma, Yunpu and Tresp, Volker and Knoll, Alois},
  booktitle={Proceedings of the AAAI Conference on Artificial Intelligence},
  volume={40},
  number={16},
  pages={13782--13790},
  year={2026}
}

@inproceedings{AutoVLA2025,
 author = {Zhou, Zewei and Cai, Tianhui and Zhao, Seth and Zhang, Yun and Huang, Zhiyu and Zhou, Bolei and Ma, Jiaqi},
 booktitle = {Advances in Neural Information Processing Systems},
 pages = {27920--27956},
 title = {AutoVLA: A Vision-Language-Action Model for End-to-End Autonomous Driving with Adaptive Reasoning and Reinforcement Fine-Tuning},
 volume = {38},
 year = {2025}
}

@article{ReasoningVLA2025,
      title={Reasoning-VLA: A Fast and General Vision-Language-Action Reasoning Model for Autonomous Driving},
      author={Dapeng Zhang and Zhenlong Yuan and Zhangquan Chen and Chih-Ting Liao and Yinda Chen and Fei Shen and Qingguo Zhou and Tat-Seng Chua},
      year={2025},
      journal={arXiv preprint arXiv:2511.19912},
}

@ARTICLE{Zheng2026PlanAgent,
  author={Zheng, Yupeng and Xing, Zebin and Zhang, Qichao and Jin, Bu and Li, Pengfei and Zheng, Yuhang and Xia, Zhongpu and Chen, Yaran and Zhao, Dongbin},
  journal={IEEE Transactions on Cognitive and Developmental Systems}, 
  title={PlanAgent: A Multi-modal Large Language Agent for Closed-loop Vehicle Motion Planning}, 
  year={2026},
  volume={},
  number={},
  pages={1-14},
  doi={10.1109/TCDS.2026.3664120}}

@inproceedings{diffusionplanner2025zheng,
title={Diffusion-Based Planning for Autonomous Driving with Flexible Guidance},
author={Yinan Zheng and Ruiming Liang and Kexin ZHENG and Jinliang Zheng and Liyuan Mao and Jianxiong Li and Weihao Gu and Rui Ai and Shengbo Eben Li and Xianyuan Zhan and Jingjing Liu},
booktitle={Proceedings of the International Conference on Learning Representations},
year={2025},
}

@inproceedings{tan2025flow,
title={Flow Matching-Based Autonomous Driving Planning with Advanced Interactive Behavior Modeling},
author={Tianyi Tan and Yinan Zheng and Ruiming Liang and Zexu Wang and Kexin Zheng and Jinliang Zheng and Jianxiong Li and Xianyuan Zhan and Jingjing Liu},
booktitle={Advances in Neural Information Processing Systems},
year={2025},
volume = {38},
pages = {38310--38335},
}

@inproceedings{diffusiondrive2025,
  title={DiffusionDrive: Truncated Diffusion Model for End-to-End Autonomous Driving},
  author={Bencheng Liao and Shaoyu Chen and Haoran Yin and Bo Jiang and Cheng Wang and Sixu Yan and Xinbang Zhang and Xiangyu Li and Ying Zhang and Qian Zhang and Xinggang Wang},
  booktitle    = {Proceedings of the IEEE/CVF Conference on Computer Vision and Pattern Recognition},
  pages        = {12037--12047},
  year         = {2025},
}

@inproceedings{Zhang2026DFP,
   author = {Zhang, Zehan and Li, Yaoyi and Zhang, Neng and Cai, Jia},
   title = {Diffusion Forcing Planner: History-Annealed Planning with Time-Dependent Guidance for Autonomous Driving},
   booktitle = {Proceedings of the IEEE/CVF Conference on Computer Vision and Pattern Recognition},
   year = {2026},
   pages = {39796-39805},
}

@InProceedings{meanfuser2026wang,
    author    = {Wang, Junli and Zheng, Yinan and Liu, Xueyi and Xing, Zebin and Li, Pengfei and Ma, Kun and Ye, Hangjun and Chen, Guang and Li, Guang and Chen, Long and Xia, Zhongpu and Zhang, Qichao},
    title     = {MeanFuser: Fast One-Step Multi-Modal Trajectory Generation and Adaptive Reconstruction via MeanFlow for End-to-End Autonomous Driving},
    booktitle = {Proceedings of the IEEE/CVF Conference on Computer Vision and Pattern Recognition},
    year      = {2026},
    pages     = {17884-17893}
}

@article{Xing2026MISTY,
      title={MISTY: High-Throughput Motion Planning via Mixer-based Single-step Drifting},
      author={Yining Xing and Zehong Ke and Yiqian Tu and Zhiyuan Liu and Wenhao Yu and Jianqiang Wang},
      year={2026},
      journal={arXiv preprint arXiv:2604.21489},
}

@inproceedings{dagger2011ross,
  title={A Reduction of Imitation Learning and Structured Prediction to No-Regret Online Learning},
  author={Ross, Stephane and Gordon, Geoffrey and Bagnell, Drew},
  booktitle = 	 {Proceedings of the International Conference on Artificial Intelligence and Statistics},
  pages = 	 {627--635},
  volume = 	 {15},
  year={2011}
}

@InProceedings{laskey2017dart,
  title={DART: Noise Injection for Robust Imitation Learning},
  author={Laskey, Michael and Lee, Jonathan and Fox, Roy and Dragan, Anca and Goldberg, Ken},
  booktitle = 	 {Proceedings of the Conference on Robot Learning},
  pages = 	 {143--156},
  year={2017},
  volume = 	 {78},
}

@InProceedings{zhang2016safedagger,
  title={Query-Efficient Imitation Learning for End-to-End Autonomous Driving},
  author={Zhang, Jiakai and Cho, Kyunghyun},
  booktitle={Proceedings of the AAAI Conference on Artificial Intelligence},
  year={2017},
  volume = 	 {30},
  pages = {2891--2897}
}

@InProceedings{kelly2018hgdagger,
  title={HG-DAgger: Interactive Imitation Learning with Human Experts},
  author={Kelly, Michael and Sidrane, Chelsea and Driggs-Campbell, Katherine and Kochenderfer, Mykel J.},
  booktitle={Proceedings of the International Conference on Robotics and Automation},
  year={2019},
 pages={8077--8083},
}

@InProceedings{menda2018ensembledagger,
  title={EnsembleDAgger: A Bayesian Approach to Safe Imitation Learning},
  author={Menda, Kunal and Driggs-Campbell, Katherine and Kochenderfer, Mykel J.},
  booktitle={Proceedings of the IEEE/RSJ International Conference on Intelligent Robots and Systems},
  year={2019},
  pages={5041--5048}
}

@InProceedings{RoaD2026garcia,
    author    = {Garcia-Cobo, Guillermo and Igl, Maximilian and Karkus, Peter and Zhang, Zhejun and Watson, Michael and Chen, Yuxiao and Ivanovic, Boris and Pavone, Marco},
    title     = {RoaD: Rollouts as Demonstrations for Closed-Loop Supervised Fine-Tuning of Autonomous Driving Policies},
    booktitle = {Proceedings of the IEEE/CVF Conference on Computer Vision and Pattern Recognition Findings},
    year      = {2026},
    pages     = {1000--1009}
}

@inproceedings{catk2025zhang,
  title = {Closed-Loop Supervised Fine-Tuning of Tokenized Traffic Models},
  author = {Zhang, Zhejun and Karkus, Peter and Igl, Maximilian and Ding, Wenhao and Chen, Yuxiao and Ivanovic, Boris and Pavone, Marco},
  booktitle = {Proceedings of the IEEE Conference on Computer Vision and Pattern Recognition},
  year = {2025},
  pages = {5422--5432}
}

@article{Wu2023HG-RL,
   author = {Wu, J. and Zhou, Y. and Yang, H. and Huang, Z. and Lv, C.},
   title = {Human-Guided Reinforcement Learning With Sim-to-Real Transfer for Autonomous Navigation},
   journal = {IEEE Transactions on Pattern Analysis and Machine Intelligence},
   volume={45},
   number={12},
   pages={14745-14759},
   ISSN = {1939-3539},
   DOI = {10.1109/TPAMI.2023.3314762},
   year = {2023},
   type = {Journal Article}
}

@INPROCEEDINGS{zhang2025carplanner,
  author={Zhang, Dongkun and Liang, Jiaming and Guo, Ke and Lu, Sha and Wang, Qi and Xiong, Rong and Miao, Zhenwei and Wang, Yue},
  booktitle={Proceedings of the IEEE/CVF Conference on Computer Vision and Pattern Recognition}, 
  title={CarPlanner: Consistent Auto-regressive Trajectory Planning for Large-scale Reinforcement Learning in Autonomous Driving}, 
  year={2025},
  pages={17239-17248},
  }

@inproceedings{Shang2025DriveDPO,
 author = {Shang, Shuyao and Chen, Yuntao and Wang, Yuqi and Li, Yingyan and ZHANG, ZHAO-XIANG},
 booktitle = {Advances in Neural Information Processing Systems},
 pages = {81565--81585},
 title = {DriveDPO: Policy Learning via Safety DPO For End-to-End Autonomous Driving},
 volume = {38},
 year = {2025}
}

@article{tang2026planr1,
      title={Plan-R1: Safe and Feasible Trajectory Planning as Language Modeling}, 
      author={Xiaolong Tang and Meina Kan and Shiguang Shan and Xilin Chen},
      year={2026},
      journal={arXiv preprint arXiv:2505.17659},
}

@article{song2026diver-rl,
      title={DIVER: Reinforced Diffusion Breaks Imitation Bottlenecks in End-to-End Autonomous Driving}, 
      author={Ziying Song and Lin Liu and Hongyu Pan and Bencheng Liao and Mingzhe Guo and Lei Yang and Yongchang Zhang and Shaoqing Xu and Caiyan Jia and Yadan Luo},
      year={2026},
      journal={arXiv preprint arXiv:2507.04049},
}

@article{zheng2026DiffusionRL,
      title={Unleashing the Potential of Diffusion Models for End-to-End Autonomous Driving}, 
      author={Yinan Zheng and Tianyi Tan and Bin Huang and Enguang Liu and Ruiming Liang and Jianlin Zhang and Jianwei Cui and Guang Chen and Kun Ma and Hangjun Ye and Long Chen and Ya-Qin Zhang and Xianyuan Zhan and Jingjing Liu},
      year={2026},
      journal={arXiv preprint arXiv:2602.22801},
}

@article{Wang2024ContinualLearningSurvey,
   author = {Wang, L. and Zhang, X. and Su, H. and Zhu, J.},
   title = {A Comprehensive Survey of Continual Learning: Theory, Method and Application},
   journal = {IEEE Transactions on Pattern Analysis and Machine Intelligence},
   volume = {46},
   number = {8},
   pages = {5362-5383},
   ISSN = {1939-3539},
   DOI = {10.1109/TPAMI.2024.3367329},
   year = {2024},
   type = {Journal Article}
}

@article{clreview2022nmi,
   author = {van de Ven, Gido M. and Tuytelaars, Tinne and Tolias, Andreas S.},
   title = {Three types of incremental learning},
   journal = {Nature Machine Intelligence},
   volume = {4},
   number = {12},
   pages = {1185-1197},
   ISSN = {2522-5839},
   DOI = {10.1038/s42256-022-00568-3},
   year = {2022},
   type = {Journal Article}
}

@inproceedings{ptmreview2024ijcai,
  title     = {Continual Learning with Pre-Trained Models: A Survey},
  author    = {Zhou, Da-Wei and Sun, Hai-Long and Ning, Jingyi and Ye, Han-Jia and Zhan, De-Chuan},
  booktitle = {Proceedings of the International Joint Conference on
               Artificial Intelligence},
  pages     = {8363--8371},
  year      = {2024},
}

@article{Kirkpatrick2017EWC,
  title={Overcoming catastrophic forgetting in neural networks},
  author={Kirkpatrick, James and Pascanu, Razvan and Rabinowitz, Neil and Veness, Joel and Desjardins, Guillaume and Rusu, Andrei A. and Milan, Kieran and Quan, John and Ramalho, Tiago and Grabska-Barwinska, Agnieszka and Hassabis, Demis and Clopath, Claudia and Kumaran, Dharshan and Hadsell, Raia},
  journal={Proceedings of the National Academy of Sciences},
  volume={114},
  number={13},
  pages={3521--3526},
  year={2017}
}

@inproceedings{Li2016LwF,
  title={Learning without Forgetting},
  author={Li, Zhizhong and Hoiem, Derek},
  booktitle={Proceedings of the European Conference on Computer Vision},
  pages={614--629},
  year={2016}
}

@inproceedings{Rebuffi2017iCaRL,
  title={iCaRL: Incremental Classifier and Representation Learning},
  author={Rebuffi, Sylvestre-Alvise and Kolesnikov, Alexander and Sperl, Georg and Lampert, Christoph H.},
  booktitle={Proceedings of the IEEE/CVF Conference on Computer Vision and Pattern Recognition },
  pages={2001--2010},
  year={2017}
}

@inproceedings{LopezPaz2017GEM,
  title={Gradient Episodic Memory for Continual Learning},
  author={Lopez-Paz, David and Ranzato, Marc'Aurelio},
  booktitle={Advances in Neural Information Processing Systems},
  year={2017},
  pages = {6470–6479}
}

@inproceedings{DEN2018Yoon,
  title={Lifelong Learning with Dynamically Expandable Networks},
  author={Yoon, Jaehong and Yang, Eunho and Lee, Jeongtae and Hwang, Sung Ju},
  booktitle={Proceedings of the International Conference on Learning Representations},
  year={2018}
}

@inproceedings{PackNet2018Mallya,
  title={PackNet: Adding Multiple Tasks to a Single Network by Iterative Pruning},
  author={Mallya, Arun and Lazebnik, Svetlana},
  booktitle={Proceedings of the IEEE/CVF Conference on Computer Vision and Pattern Recognition},
  pages={7765--7773},
  year={2018}
}

@inproceedings{Wang2022L2P,
  title={Learning to Prompt for Continual Learning},
  author={Wang, Zifeng and Zhang, Zizhao and Ebrahimi, Sayna and Sun, Ruoxi and Zhang, Han and Lee, Chen-Yu and Ren, Xiaoqi and Su, Guolong and Perot, Vincent and Dy, Jennifer and Pfister, Tomas},
  booktitle={Proceedings of the IEEE/CVF Conference on Computer Vision and Pattern Recognition},
  pages={139--149},
  year={2022}
}

@inproceedings{Wang2022DualPrompt,
  title={DualPrompt: Complementary Prompting for Rehearsal-free Continual Learning},
  author={Wang, Zifeng and Zhang, Zizhao and Lee, Chen-Yu and Zhang, Han and Sun, Ruoxi and Ren, Xiaoqi and Su, Guolong and Perot, Vincent and Dy, Jennifer and Pfister, Tomas},
  booktitle={Proceedings of the European Conference on Computer Vision},
  pages={631--648},
  year={2022}
}

@INPROCEEDINGS{inflora2024liang,
  author={Liang, Yan-Shuo and Li, Wu-Jun},
  booktitle={Proceedings of the IEEE/CVF Conference on Computer Vision and Pattern Recognition}, 
  title={InfLoRA: Interference-Free Low-Rank Adaptation for Continual Learning}, 
  year={2024},
  volume={},
  number={},
  pages={23638-23647},
  }

@article{cdlreview2023li,
   author = {Li, Zirui and Gong, Cheng and Lin, Yunlong and Li, Guopeng and Wang, Xinwei and Lu, Chao and Wang, Miao and Chen, Shanzhi and Gong, Jianwei},
   title = {Continual driver behaviour learning for connected vehicles and intelligent transportation systems: Framework, survey and challenges},
   journal = {Green Energy and Intelligent Transportation},
   volume = {2},
   number = {4},
   ISSN = {27731537},
   DOI = {10.1016/j.geits.2023.100103},
   year = {2023},
   pages = {100103},
}

@article{Li2026ContinualLearningInteractiveTrajectoryPrediction, 
author = {Huiqian Li and Xiaozhou Wu and Jin Huang and Zhihua Zhong},
title = {Toward zero-forget continual learning for interactive trajectory prediction: A dynamically expandable approach},
year = {2026},
journal = {Communications in Transportation Research},
volume = {6},
number = {1},
pages = {9640015},
doi = {10.26599/COMMTR.2026.9640015},
}

@ARTICLE{Lin2026H2CPrediction,
  author={Lin, Yunlong and Li, Zirui and Du, Guodong and Zhao, Xiaocong and Gong, Cheng and Wang, Xinwei and Lu, Chao and Gong, Jianwei},
  journal={IEEE Transactions on Intelligent Transportation Systems}, 
  title={H2C: Hippocampal Circuit-Inspired Continual Learning for Lifelong Trajectory Prediction in Autonomous Driving}, 
  year={2026},
  volume={},
  number={},
  pages={1-18},
  doi={10.1109/TITS.2026.3679635}}

@article{Cao2023ContinuousImprovement,
   author = {Cao, Zhong and Jiang, Kun and Zhou, Weitao and Xu, Shaobing and Peng, Huei and Yang, Diange},
   title = {Continuous improvement of self-driving cars using dynamic confidence-aware reinforcement learning},
   journal = {Nature Machine Intelligence},
   volume = {5},
   number = {2},
   pages = {145-158},
   ISSN = {2522-5839},
   DOI = {10.1038/s42256-023-00610-y},
   year = {2023},
}

@article{Meng2025PreservingKnowledgeCombining,
   author = {Meng, Yuan and Bing, Zhenshan and Yao, Xiangtong and Chen, Kejia and Huang, Kai and Gao, Yang and Sun, Fuchun and Knoll, Alois},
   title = {Preserving and combining knowledge in robotic lifelong reinforcement learning},
   journal = {Nature Machine Intelligence},
   volume = {7},
   number = {2},
   pages = {256-269},
   ISSN = {2522-5839},
   DOI = {10.1038/s42256-025-00983-2},
   year = {2025},
}

@article{Gong2024BeyondImitationLifeLongPolicyLearning,
   author = {Gong, C. and Lu, C. and Li, Z. and Liu, Z. and Gong, J. and Chen, X.},
   title = {Beyond Imitation: A Life-long Policy Learning Framework for Path Tracking Control of Autonomous Driving},
   journal = {IEEE Transactions on Vehicular Technology},
   volume = {73},
   number = {7},
   pages = {9786-9799},
   ISSN = {1939-9359},
   DOI = {10.1109/TVT.2024.3382309},
   year = {2024},
}

@article{Yang2025HGCContinualLearning,
   author = {Yang, H. and Zhou, Y. and Wu, J. and Liu, H. and Yang, L. and Lv, C.},
   title = {Human-Guided Continual Learning for Personalized Decision-Making of Autonomous Driving},
   journal = {IEEE Transactions on Intelligent Transportation Systems},
   year={2025},
   volume={26},
   number={4},
   pages={5435-5447},
   DOI = {10.1109/TITS.2024.3524609},
}

@inproceedings{Buzzega2020DER,
 author = {Buzzega, Pietro and Boschini, Matteo and Porrello, Angelo and Abati, Davide and CALDERARA, SIMONE},
 booktitle = {Advances in Neural Information Processing Systems},
 pages = {15920--15930},
 title = {Dark Experience for General Continual Learning: a Strong, Simple Baseline},
 volume = {33},
 year = {2020}
}

@INPROCEEDINGS{Karnchanachari2024NuplanBenchmark,
   author = {Karnchanachari, Napat and Geromichalos, Dimitris and Tan, Kok Seang and Li, Nanxiang and Eriksen, Christopher and Yaghoubi, Shakiba and Mehdipour, Noushin and Bernasconi, Gianmarco and Fong, Whye Kit and Guo, Yiluan and Caesar, Holger},
   title = {Towards learning-based planning: The nuPlan benchmark for real-world autonomous driving},
   booktitle={Proceedings of the IEEE International Conference on Robotics and Automation}, 
   year = {2024},
   pages = {629--636}
}

\end{document}